# Towards Adjustable Autonomy for the Real World

**Paul Scerri**                                                    SCERRI@ISI.EDU
**David V. Pynadath**                                        PYNADATH@ISI.EDU
**Milind Tambe**                                                TAMBE@USC.EDU
*Information Sciences Institute and Computer Science Department*
*University of Southern California*
*4676 Admiralty Way, Marina del Rey, CA 90292 USA*

## Abstract

Adjustable autonomy refers to entities dynamically varying their own autonomy, transferring decision-making control to other entities (typically agents transferring control to human users) in key situations. Determining whether and when such transfers-of-control should occur is arguably the fundamental research problem in adjustable autonomy. Previous work has investigated various approaches to addressing this problem but has often focused on individual agent-human interactions. Unfortunately, domains requiring collaboration between teams of agents and humans reveal two key shortcomings of these previous approaches. First, these approaches use rigid one-shot transfers of control that can result in unacceptable coordination failures in multiagent settings. Second, they ignore costs (e.g., in terms of time delays or effects on actions) to an agent's team due to such transfers-of-control.

To remedy these problems, this article presents a novel approach to adjustable autonomy, based on the notion of a *transfer-of-control strategy*. A transfer-of-control strategy consists of a conditional sequence of two types of actions: (i) actions to transfer decision-making control (e.g., from an agent to a user or vice versa) and (ii) actions to change an agent's pre-specified coordination constraints with team members, aimed at minimizing miscoordination costs. The goal is for high-quality individual decisions to be made with minimal disruption to the coordination of the team. We present a mathematical model of transfer-of-control strategies. The model guides and informs the operationalization of the strategies using Markov Decision Processes, which select an optimal strategy, given an uncertain environment and costs to the individuals and teams. The approach has been carefully evaluated, including via its use in a real-world, deployed multi-agent system that assists a research group in its daily activities.

## 1. Introduction

Exciting, emerging application areas ranging from intelligent homes (Lesser et al., 1999), to routine organizational coordination (Pynadath et al., 2000), to electronic commerce (Collins et al., 2000a), to long-term space missions (Dorais et al., 1998) utilize the decision-making skills of both agents and humans. These new applications have brought forth an increasing interest in agents' *adjustable autonomy* (AA), i.e., in entities *dynamically adjusting their own level of autonomy based on the situation* (Mulsiner & Pell, 1999). Many of these exciting applications will not be deployed unless reliable AA reasoning is a central component. With AA, an entity need not make all decisions autonomously; rather it can choose to reduce its own autonomy and *transfer decision-making control* to other users or agents, when doing so is expected to have some net benefit (Dorais et al., 1998; Barber, Goel, & Martin, 2000a; Hexmoor & Kortenkamp, 2000).

A central problem in AA is to determine whether and when transfers of decision-making control should occur. A key challenge is to balance two potentially conflicting goals. On the one hand, to ensure that the highest-quality decisions are made, an agent can transfer control to a human user (or another agent) whenever that user has superior decision-making expertise.[1] On the other hand, interrupting a user has high costs and the user may be unable to make and communicate a decision, thus such transfers-of-control should be minimized. Previous work has examined several different techniques that attempt to balance these two conflicting goals and thus address the transfer-of-control problem. For example, one technique suggests that decision-making control should be transferred if the expected utility of doing so is higher than the expected utility of making an autonomous decision (Horvitz, Jacobs, & Hovel, 1999). A second technique uses uncertainty as the sole rationale for deciding who should have control, forcing the agent to relinquish control to the user whenever uncertainty is high (Gunderson & Martin, 1999). Yet other techniques transfer control to a user if an erroneous autonomous decision could cause significant harm (Dorais et al., 1998) or if the agent lacks the capability to make the decision (Ferguson, Allen, & Miller, 1996).

Unfortunately, these previous approaches to transfer-of-control reasoning and indeed most previous work in AA, have focused on domains involving a single agent and a single user, isolated from interactions with other entities. When applied to interacting teams of agents and humans, where interaction between an agent and a human impacts the interaction with other entities, these techniques can lead to dramatic failures. In particular, the presence of other entities as team members introduces a third goal of maintaining coordination (in addition to the two goals already mentioned above), which these previous techniques fail to address. Failures occur for two reasons. Firstly, these previous techniques ignore team related factors, such as costs to the team due to incorrect decisions or due to delays in decisions during such transfers-of-control. Secondly (and more importantly), these techniques use one-shot transfers-of-control, rigidly committing to one of two choices: (i) transfer control and wait for input (choice $H$) or (ii) act autonomously (choice $A$). However, given interacting teams of agents and humans, either choice can lead to costly failures if the entity with control fails to make or report a decision in a way that maintains coordination. For instance, a human user might be unable to provide the required input due to a temporary communication failure; this may cause an agent to fail in its part of a joint action, as this joint action may be dependent on the user's input. On the other hand, forcing a less capable entity to make a decision simply to avoid miscoordination can lead to poor decisions with significant consequences. Indeed, as seen in Section 2.2, when we applied a rigid transfer-of-control decision-making to a domain involving teams of agents and users, it failed dramatically.

Yet, many emerging applications do involve multiple agents and multiple humans acting cooperatively towards joint goals. To address the shortcomings of previous AA work in such domains, this article introduces the notion of a *transfer-of-control strategy*. A transfer-of-control strategy consists of a pre-defined, conditional sequence of two types of actions: (i) actions to transfer decision-making control (e.g., from an agent to a user or vice versa); (ii) actions to change an agent's pre-specified coordination constraints with team members, rearranging activities as needed (e.g., reordering tasks to buy time to make the decision). The agent executes such a strategy by performing the actions in order, transferring control to the specified entity and changing coordination as required, until some point in time when the entity currently in control exercises that control and makes the decision. Thus, the previous choices of $H$ or $A$ are just two of many different and possibly more complex transfer-of-control strategies. For instance, an $A\mathcal{D}AH$ strategy implies that an agent initially attempts to make an autonomous decision. If the agent makes the decision autonomously the strategy execution ends there. However, there is a chance that it is unable to make the decision in a timely manner, perhaps because its computational resources are busy with higher priority tasks. To avoid miscoordination the agent executes a $\mathcal{D}$ action which changes the coordination constraints on the activity. For example, a $\mathcal{D}$ action could be to inform other agents that the coordinated action will be delayed, thus incurring a cost of inconvenience to others but buying more time to make the decision. If it still cannot make the decision, it will eventually take action $H$, transferring decision-making control to the user and waiting for a response. In general, strategies can involve all available entities and contain many actions to change coordination constraints. While such strategies may be useful in single-agent single-human settings, they are particularly critical in general multiagent settings, as discussed below.

---

1. While the AA problem in general involves transferring control from one entity to another, in this paper, we will typically focus on interactions involving autonomous agents and human users.

Transfer-of-control strategies provide a flexible approach to AA in complex systems with many actors. By enabling multiple transfers-of-control between two (or more) entities, rather than rigidly committing to one entity (i.e., $A$ or $H$), a strategy attempts to provide the highest quality decision, while avoiding coordination failures. In particular, in a multiagent setting there is often uncertainty about whether an entity will make a decision and when it will do so, e.g., a user may fail to respond, an agent may not be able to make a decision as expected or a communication channel may fail. A strategy addresses such uncertainty by planning multiple transfers of control to cover for such contingencies. For instance, with the $A\mathcal{D}H$ strategy, an agent ultimately transfers control to a human to attempt to ensure that some response will be provided in case the agent is unable to act. Furthermore, explicit coordination-change actions, i.e., $\mathcal{D}$ actions, reduce miscoordination effects, for a cost, while better decisions are being made. Finally, since the utility of transferring control or changing coordination is dependent on the actions taken afterwards, the agent must plan a strategy in advance to find the sequence of actions that maximizes team benefits. For example, reacting to the current situation and repeatedly taking and giving control as in the strategy $A\mathcal{D}HA\mathcal{D}H \ldots$ may be more costly than planning ahead, making a bigger coordination change, and using a shorter $A\mathcal{D}H$ strategy. We have developed a decision theoretic model of such strategies, that allows the expected utility of a strategy to be calculated and, hence, strategies to be compared.

Thus, a key AA problem is to select the right strategy, i.e., one that provides the benefit of high-quality decisions without risking significant costs in interrupting the user and miscoordination with the team. Furthermore, an agent must select the right strategy despite significant uncertainty. Markov decision processes (MDPs) (Puterman, 1994) are a natural choice for implementing such reasoning because they explicitly represent costs, benefits and uncertainty as well as doing lookahead to examine the potential consequences of sequences of actions. In Section 4, a general reward function is presented for an MDP that results in an agent carefully balancing risks of incorrect autonomous decisions, potential miscoordination and costs due to changing coordination between team members. Detailed experiments were performed on the MDP, the key results of which are as follows. As the relative importance of central factors, such as the cost of miscoordination, was varied the resulting MDP policies varied in a desirable way, i.e., the agent made more decisions autonomously if the cost of transferring control to other entities increased. Other experiments reveal a phenomenon not reported before in the literature: an agent may act more autonomously when coordination change costs are either too low or too high, but in a "middle" range, the agent tends to act less autonomously.

Our research has been conducted in the context of a real-world multi-agent system, called *Electric Elves* (E-Elves) (Chalupsky, Gil, Knoblock, Lerman, Oh, Pynadath, Russ, & Tambe, 2001; Pynadath et al., 2000), that we have used for over six months at the University of Southern California, Information Sciences Institute. The E-Elves assists a group of researchers and a project assistant in their daily activities, providing an exciting opportunity to test AA ideas in a real environment. Individual user proxy agents called Friday (from Robinson Crusoe's servant Friday) act in a team to assist with rescheduling meetings, ordering meals, finding presenters and other day-to-day activities. Over the course of several months, MDP-based AA reasoning was used around the clock in the E-Elves, making many thousands of autonomy decisions. Despite the unpredictability of the user's behavior and the agent's limited sensing abilities, the MDP consistently made sensible AA decisions. Moreover, many times the agent performed several transfers-of-control to cope with contingencies such as a user not responding. One lesson learned when actually deploying the system was that sometimes users wished to influence the AA reasoning, e.g., to ensure that control was transferred to them in particular circumstances. To allow users to influence the AA reasoning, safety constraints are introduced that allow users to prevent agents from taking particular actions or ensuring that they do take particular actions. These safety constraints provide guarantees on the behavior of the AA reasoning, making the basic approach more generally applicable and, in particular, making it more applicable to domains where mistakes have serious consequences.

The rest of this article is organized as follows. Section 2 gives a detailed description of the AA problem and presents the Electric Elves as a motivating example application. Section 3 presents a formal model of transfer-of-control strategies for AA. (Readers not interested in the mathematical details may wish to skip over Section 3.) The operationalization of the strategies via MDPs is described in Section 4. In Section 5, the results of detailed experiments are presented. Section 6 looks at related work, including how earlier AA work can be analyzed within the strategies framework. Section 7 gives a summary of the article. Finally, Section 8 outlines areas where the work could be extended to make it applicable to more applications.

## 2. Adjustable Autonomy – The Problem

The general AA problem has not been previously formally defined in the literature, particularly for a multiagent context. In the following, a formal definition of the problem is given so as to clearly define the task for the AA reasoning. The team, which may consist entirely of agents or include humans, has some joint activity, $\alpha$. Each entity in the team works cooperatively on the joint activity. The agent, $A$, has a role, $\rho$, in the team. Depending on the specific task, some or all of the roles will need to be performed successfully in order for the joint activity to succeed. The primary goal of the agent is the success of $\alpha$ which it pursues by performing $\rho$. Performing $\rho$ requires that one or more non-trivial decisions are made. To make a decision, $d$, the agent can draw upon $n$ other entities from a set $E = \{e_1 \ldots e_n\}$, which typically includes the agent itself. Each entity in $E$ (e.g., a human user) is capable of making decision $d$. The entities in $E$ are not necessarily part of the team performing $\alpha$. Different agents and users will have differing abilities to make decisions due to available computational resources, access to relevant information, etc. Coordination *constraints*, $\asymp$, exist between $\rho$ and the roles of other members of the team. For example, various roles might need to be executed simultaneously or in a certain order or with some combined quality or total cost. A critical facet of the successful completion of the joint task $\alpha$, given its jointness, is to ensure that coordination between team members is maintained, i.e., $\asymp$ are not violated. Thus, we can describe an AA problem instance with the tuple: $\langle A, \alpha, \rho, \asymp, d, E \rangle$.

From an AA perspective, the agent can take two types of actions for a decision, $d$. First, it can transfer control to an entity in $E$ capable of making that decision. In general, there are no restrictions on when, how often or for how long decision-making control can be transferred to a particular entity. Typically, the agent can also transfer decision-making control to itself. In general, we assume that when the agent transfers control, it does not have any guarantee on the exact time of response or exact quality of the decision made by the entity to which control is transferred. In fact, in some cases it will not know whether the entity will be able to make a decision at all or even whether the entity will know it has decision-making control, e.g., if control was transferred via email, the agent may not know if the user actually read the email.

The second type of action that an agent can take is to request changes in the coordination constraints, $\asymp$, between team members. A coordination change gives the agent the possibility of changing the requirements surrounding the decision to be made, e.g., the required timing, cost or quality of the decision, which may allow it to better fulfill its responsibilities. A coordination change might involve reordering or delaying tasks or it may involve changing roles, or it may be a more dramatic change where the team pursues $\alpha$ in a completely different way. Changing coordination has some cost, but it may be better to incur this cost than violate coordination constraints, i.e., incur miscoordination costs. Miscoordination between team members will occur for many reasons, e.g., a constraint that limits the total cost of a joint task might be violated if one team member incurs a higher than expected cost and other team members do not reduce their costs. In this article, we are primarily concerned with constraints related to the timing of roles, e.g., ordering constraints or requirements on simultaneous execution. This in turn, usually requires that the agent guards against delayed decisions although it can also require that a decision is not made too soon.

Thus, the AA problem for the agent, given a problem instance, $\langle A, \alpha, \rho, \asymp, d, E \rangle$, is to choose the transfer-of-control or coordination-change actions that maximizes the overall expected utility of the team. In the remainder of this section we describe a concrete, real-world domain for AA (Section 2.1) and an initial failed approach that motivates our solution (Section 2.2).

## 2.1 The Electric Elves

This research was initiated in response to issues that arose in a real application and the resulting approach was extensively tested in the day-to-day running of that application. The Electric Elves (E-Elves) is a project at USC/ISI to deploy an agent organization in support of the daily activities of a human organization (Pynadath et al., 2000; Chalupsky et al., 2001). We believe this application to be fairly typical of future generation applications involving teams of agents and humans. The operation of a human organization requires the performance of many everyday tasks to ensure coherence in organizational activities, e.g., monitoring the status of activities, gathering information and keeping everyone informed of changes in activities. Teams of software agents can aid organizations in accomplishing these tasks, facilitating coherent functioning and rapid, flexible response to crises. A number of underlying AI technologies support the E-Elves, e.g., technologies devoted to agent-human interactions, agent coordination, accessing multiple heterogeneous information sources, dynamic assignment of organizational tasks, and deriving information about organization members (Chalupsky et al., 2001). While these technologies are useful, AA is fundamental to the effective integration of the E-Elves into the day-to-day running of a real organization and, hence, is the focus of this paper.

The basic design of the E-Elves is shown in Figure 1(a). Each agent proxy is called Friday (after Robinson Crusoes' man-servant Friday) and acts on behalf of its user in the agent team. The design of the Friday proxies is discussed in detail in (Tambe, Pynadath, Chauvat, Das, & Kaminka, 2000) (where they are referred to as TEAMCORE proxies). Currently, Friday can perform several tasks for its user. If a user is delayed to a meeting, Friday can reschedule the meeting, informing other Fridays, who in turn inform their users. If there is a research presentation slot open, Friday may respond to the invitation to present on behalf of its user. Friday can also order its user's meals (see Figure 2(a)) and track the user's location, posting it on a Web page. Friday communicates with users using wireless devices, such as personal digital assistants (PALM VIIs) and WAP-enabled mobile phones, and via user workstations. Figure 1(b) shows a PALM VII connected to a Global Positioning Service (GPS) device, for tracking users' locations and enabling wireless communication between Friday and a user. Each Friday's team behavior is based on a teamwork model, called STEAM (Tambe, 1997). STEAM encodes and enforces the constraints between roles that are required for the success of the joint activity, e.g., meeting attendees should arrive at a meeting simultaneously. When a role within the team needs to be filled, STEAM requires that a team member is assigned responsibility for that role. To find the best suited person, the team auctions off the role, allowing it to consider a combination of factors and assign the best suited user. Friday can bid on behalf of its user, indicating whether its user is capable and/or willing to fill a particular role. Figure 2(b) shows a tool that allows users to view auctions in progress and intervene if they so desire. In the auction in progress, Jay Modi's Friday has bid that Jay is capable of giving the presentation, but is unwilling to do so. Paul Scerri's agent has the highest bid and was eventually allocated the role.

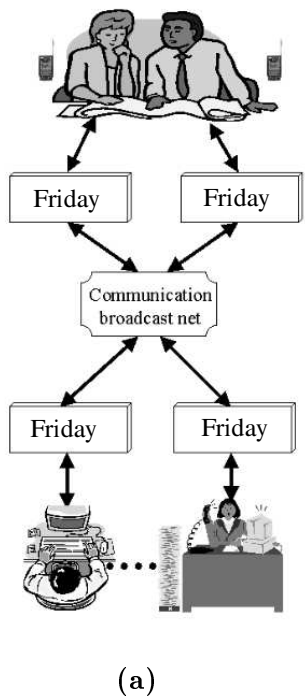

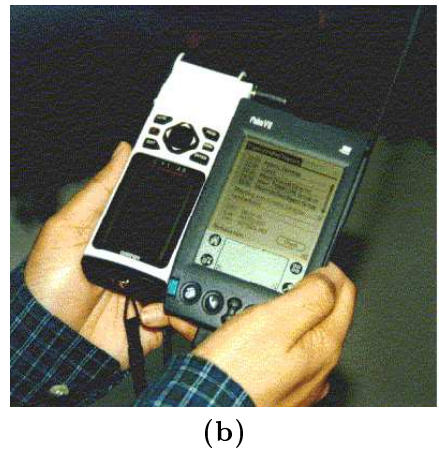

(a)                                    (b)

Figure 1: (a) Overall E-Elves architecture, showing Friday agents interacting with users. (b)Palm VII for communicating with users and GPS device for detecting their location.

AA is critical to the success of the E-Elves since, despite the range of sensing devices, Friday has considerable uncertainty about the user's intentions and even location; hence, Friday will not always have the appropriate information to make correct decisions. On the other hand, while the user has the required information, Friday cannot continually ask the user for input, since such interruptions are disruptive and time-consuming. There are four decisions in the E-Elves to which AA reasoning is applied: (i) whether a user will attend a meeting on time; (ii) whether to close an auction for a role; (iii) whether the user is willing to perform an open team role; and (iv) if and what to order for lunch. In this paper, we focus on the AA reasoning for two of those decisions: whether a user will attend a meeting on time and whether to close an auction for a role. The decision as to whether a user will attend a meeting on time is the most often used and most difficult of the decisions Friday faces. We briefly describe the decision to close an auction and later show how an insight provided by the model of strategies led to a significant reduction in the amount of code required to implement the AA reasoning for that decision. The decision to volunteer a user for a meeting is similar to the earlier decisions, and omitted for brevity; the decision to order lunch is currently implemented in a simpler fashion and is not (at least as yet) illustrative of the full set of complexities.

A central decision for Friday, which we describe in terms of our problem formulation, $\langle A, \alpha, \rho, \asymp, d, E \rangle$, is whether its user will attend a meeting at the currently scheduled meeting time. In this case, Friday is the agent, $A$. The joint activity, $\alpha$, is for the meeting attendees to attend the meeting simultaneously. Friday acts as proxy for its user, hence its role, $\rho$, is to ensure that its user arrives at the currently scheduled meeting time. The coordination constraint, $\asymp$, between Friday's role and the roles of other Fridays is that they occur simultaneously, i.e., the users must attend at the currently scheduled time. If any attendee arrives late, or not at all, the time of all the attendees is wasted; on the other hand, delaying a meeting is disruptive to users' schedules. The decision, $d$, is whether the user will attend the meeting or not and could be made by either Friday or the user, i.e., $E = \{\text{user}, \text{Friday}\}$. Clearly, the user will be often better placed to make this decision. However, if Friday transfers control to the user for the decision, it must guard against miscoordination, i.e., having the other attendees wait, while waiting for a user response. Some decisions are potentially costly, e.g., incorrectly telling the other attendees that the user will not attend, and Friday should avoid taking them autonomously. To buy more time for the user to make the decision or for itself to gather more information, Friday could change coordination constraints with a $\mathcal{D}$ action. Friday has several different $\mathcal{D}$ actions at its disposal, including delaying the meeting by different lengths of time, as well as being able to cancel the meeting entirely. The user can also request a $\mathcal{D}$ action, e.g., via the dialog box in Figure 5(a), to buy more time to make it to the meeting. If the user decides a $\mathcal{D}$ is required, Friday is the conduit through which other Fridays (and hence their users) are informed. Friday must select a sequence of actions, either transferring control to the user, delaying or cancelling the meeting or autonomously announcing that the user will or will not attend, to maximize the utility of the team.

The second AA decision that we look at is the decision to close an auction for an open role and assign a user to that role.[2] In this case, the joint activity, $\alpha$, is the group research meeting and the role, $\rho$, is to be the auctioneer. Users will not always submit bids for the role immediately; in fact, the bids may be spread out over several days, or some users might not bid at all. The specific decision, $d$, on which we focus is whether to close the auction and assign the role or continue waiting for incoming bids. Once individual team members provide their bids, the auctioneer agent or human team leader decides on a presenter based on that input ($E = \{$user, auctioneer agent$\}$). The team expects a willing presenter to do a high-quality research presentation, which means the presenter will need some time to prepare. Thus, the coordination constraint, $\asymp$ is that the most capable, willing user must be allocated to the role with enough time to prepare the presentation. Despite individually responsible actions, the agent team may reach a highly undesirable decision, e.g., assigning the same user week after week, hence there is advantage in getting the human team leader's input. The agent faces uncertainty (e.g., will better bids come in?), costs (i.e., the later the assignment, the less time the presenter has to prepare), and needs to consider the possibility that the human team leader has some special preference about who should do a presentation at some particular meeting. By transferring control, the agent allows the human team leader to make an assignment. For this decision, a coordination-change action, $\mathcal{D}$, would reschedule the research meeting. However, relative to the cost of cancelling the meeting, the cost of rescheduling is too high for rescheduling to be a useful action.

---

2. There are also roles for submitting bids to the auction but the AA for those decisions is simpler, hence we do not focus on them here.

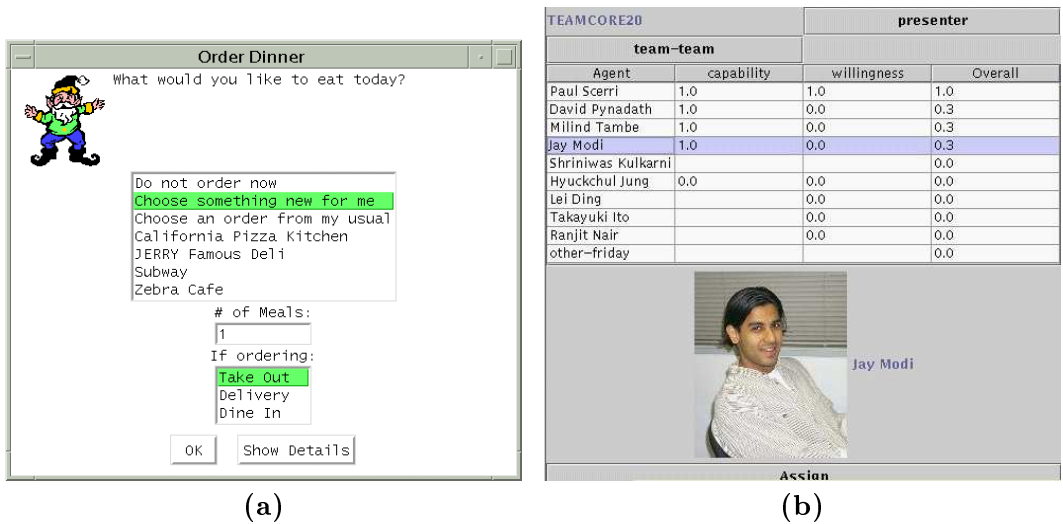

Figure 2: (a) Friday transferring control to the user for a decision whether to order lunch. (b) The E-Elves auction monitoring tool.

## 2.2 Decision-Tree Approach

One logical avenue of attack on the AA problem for the E-Elves was to apply an approach used in a previously reported, successful meeting scheduling system, in particular CAP (Mitchell, Caruana, Freitag, McDermott, & Zabowski, 1994). Like CAP, Friday learned user preferences using C4.5 decision-tree learning (Quinlan, 1993). Friday recorded values of a dozen carefully selected attributes and the user's preferred action (identified by asking the user) whenever it had to make a decision. Friday used the data to learn a decision tree that encoded its autonomous decision making. For AA, Friday also asked if the user wanted such decisions taken autonomously in the future. From these responses, Friday used C4.5 to learn a second decision tree which encoded its rules for transferring control. Thus, if the second decision tree indicated that Friday should act autonomously, it would take the action suggested by the first decision tree. Initial tests with the C4.5 approach were promising (Tambe et al., 2000), but a key problem soon became apparent. When Friday encountered a decision for which it had learned to transfer control to the user, it would wait indefinitely for the user to make the decision, even though this inaction caused miscoordination with teammates. In particular, other team members would arrive at the meeting location, waiting for a response from the user's Friday, but they would end up completely wasting their time as no response arrived. To address this problem, if a user did not respond within a fixed time limit (five minutes), Friday took an autonomous action. Although performance improved, when the resulting system was deployed 24/7 it led to some dramatic failures, including:

1. Example 1: Tambe's (a user) Friday incorrectly cancelled a meeting with the division director because Friday over-generalized from training examples.

2. Example 2: Pynadath's (another user) Friday incorrectly cancelled the group's weekly research meeting when a time-out forced the choice of an (incorrect) autonomous action.

3. Example 3: A Friday delayed a meeting almost 50 times, each time by 5 minutes. It was correctly applying a learned rule but ignoring the nuisance to the rest of the meeting participants.

4. Example 4: Tambe's Friday automatically volunteered him for a presentation, but he was actually unwilling. Again Friday over-generalized from a few examples and when a timeout occurred it took an undesirable autonomous action.

Clearly, in a team context, rigidly transferring control to one agent (user) failed. Furthermore, using a time-out that rigidly transferred control back to the agent, when it was not capable of making a high-quality decision, also failed. In particular, the agent needed to better avoid taking risky decisions by explicitly considering their costs (example 1), or take lower cost actions to delay meetings to buy the user more time to respond (example 2 and 4). Furthermore, as example 3 showed, the agent needed to plan ahead, to avoid taking costly sequences of actions that could be replaced by a single less costly action (example 3). In theory, using C4.5 Friday might have eventually been able to learn rules that would successfully balance costs and deal with uncertainty and handle all the special cases and so on, but a very large amount of training data would be required.

## 3. Strategies for Adjustable Autonomy

To avoid rigid one-shot transfers of control and allow team costs to be considered, we introduce the notion of a *transfer-of-control strategy*, which is defined as follows:

**Definition 3.1** *A transfer-of-control strategy is a pre-defined, conditional sequence of two types of actions: (i) actions to transfer decision-making control (e.g., from an agent to a user or other agents, or vice versa) and (ii) actions to change an agent's pre-specified coordination constraints with team members, aimed at minimizing miscoordination costs.*

The agent executes a transfer-of-control strategy by performing the specified actions in sequence, transferring control to the specified entity and changing coordination as required, until some point in time when the entity currently in control exercises that control and makes the decision. Considering multi-step strategies allows an agent to exploit decision-making sources considered too risky to exploit without the possibility of retaking control. For example, control could be transferred to a very capable but not always available decision maker then taken back if the decision was not made before serious miscoordination occurred. More complex strategies, potentially involving several coordination changes, give the agent the option to try several decision-making sources or to be more flexible in getting input from high-quality decision makers. As a result, transfer-of-control strategies specifically allow an agent to avoid costly errors, such as those enumerated in the previous section.[3]

Given an AA problem instance, $\langle A, \alpha, \rho, \asymp, d, E \rangle$, agent $A$ can transfer decision-making control for a decision $d$ to any entity $e_i \in E$, and we denote such a transfer-of-control action with the symbol representing the entity, i.e., transferring control to $e_i$ is denoted as $e_i$. When the agent transfers decision-making control, it may stipulate a limit on the time that it will wait for a response from that entity. To capture this additional stipulation, we denote transfer-of-control actions with this time limit, e.g., $e_i(t)$ represents that $e_i$ has decision-making control for a maximum time of $t$. Such an action has two possible outcomes: either $e_i$ responds before time $t$ and makes the decision, or it does not respond and decision $d$ remains unmade at time $t$. In addition, the agent has some mechanism by which it can change coordination constraints (denoted $\mathcal{D}$) to change the expected timing of the decision. The $\mathcal{D}$ action changes the coordination constraints, $\asymp$, between team members. The action has an associated value, $\mathcal{D}_{value}$, which specifies its magnitude (i.e., how much the $\mathcal{D}$ has alleviated the temporal pressure), and a cost, $\mathcal{D}_{cost}$, which specifies the price paid for making the change. We can concatenate such actions to specify a complete transfer-of-control strategy. For instance, the strategy $H(5)A$ would specify that the agent first relinquishes control and asks entity $H$ (denoting the *H*uman user). If the user responds with a decision within five minutes, then there is no need to go further. If not, then the agent proceeds to the next transfer-of-control action in the sequence. In this example, this next action, $A$, specifies that the agent itself make the decision and complete the task. No further transfers of control occur in this case. We can define the space of all possible strategies with the following regular expression:

$$\mathbf{S} = (E \times \mathcal{R})((E \times \mathcal{R}) + \mathcal{D}) * \tag{1}$$

where $(E \times \mathcal{R})$ is all possible combinations of entity and maximum time.

For readability, we will frequently omit the time specifications from the transfer-of-control actions and instead write just the order in which the agent transfers control among the entities and executes $\mathcal{D}$s (e.g., we will often write $HA$ instead of $H(5)A$). If time specifications are omitted, we assume the transfers happen at the optimal times,[4] i.e., the times that lead to highest expected utility. If we consider strategies with the same sequence of actions but different timings to be the same strategy, the agent has $O(|E|^k)$ possible strategies to select from, where $k$ is the maximum length of the strategy and $|E|$ is the number of entities. Thus, the agent has a wide range of options, even if practical considerations lead to a reasonable upper bound on $k$ and $|E|$. The agent must select the strategy that maximizes the overall expected utility of $\alpha$.

---

3. In some domains, it may make sense to attempt to get input from more than one entity at once, hence requiring strategies that have actions that might be executed in parallel. However, in this work, as a first step, we do not consider such strategies. Furthermore, they are not relevant for the domains at hand.

In the rest of this section, we present a mathematical model of transfer-of-control strategies for AA and use that model to guide the search for a solution. Moreover, the model provides a tool for predicting the performance of various strategies, justifying their use and explaining observed phenomena of their use. Section 3.1 presents the model of AA strategies in detail. Section 3.2 reveals key properties of complex strategies, including dominance relationships among strategies. Section 3.3 examines the E-Elves application in the light of the model, to make specific predictions about some properties that a successful AA approach reasoning for that application class will have. These predictions shape the operationalization of strategies in Section 4.

## 3.1 A Mathematical Model of Strategies

The transfer-of-control model presented in this section allows calculation of the expected utility (EU) of individual strategies, thus allowing strategies to be compared. The calculation of a strategy's EU considers four elements: the likely relative quality of different entities' decisions; the probability of getting a response from an entity at a particular time; the cost of delaying a decision; and the costs and benefits of changing coordination constraints. While other parameters might also be modeled in a similar manner, our experience with the E-Elves and other AA work suggests that these parameters are the critical ones across a wide range of joint activities.

The first element of the model is the expected quality of an entity's decision. In general, we capture the quality of an entity's decision at time $t$ with the functions $\mathbf{EQ} = \{EQ_e^d(t) : \mathcal{R} \to \mathcal{R}\}$. The quality of a decision reflects both the probability that the entity will make an "appropriate" decision and the costs incurred if the decision is wrong. The expected quality of a decision is calculated in a decision theoretic way, by multiplying the probability of each outcome, i.e., each decision, by the utility of that decision, i.e., the cost or benefit of that decision. For example, the higher the probability that the entity will make a mistake, the lower the quality, even lower if the mistakes might be very costly. The quality of decision an entity will make can vary over time as the information available to it changes or as it has more time to "think". The second element of the model is the probability that an entity will make a decision if control is transferred to it. The functions $\mathbf{P} = \{P_\top^e(t) : \mathcal{R} \to [0,1]\}$, represent continuous probability distributions over the time that the entity $e$ will respond. That is, the probability that $e_i$ will respond before time $t_0$ is $\int_0^{t_0} P_\top^{e_i}(t)dt$.

The third element of the model is a representation of the cost of inappropriate timing of a decision. In general, not making a decision until a particular point in time incurs some cost that is a function of both the time, $t$, and the coordination constraints, $\asymp$, between team members. As stated earlier, we focus on cases of constraint violations due to delays in making decisions. Thus, the cost is due to the violation of the constraints caused by not making a decision until that point in time. We can write down a *wait-cost function*: $\mathbf{W} = f(\asymp, t)$ which returns the cost of not making a decision until a particular point in time given coordination constraints, $\asymp$. This miscoordination cost is a fundamental aspect of our model given our emphasis on multiagent domains. It is called a "wait cost" because it models the miscoordination that arises while the team "waits" for some entity to make the ultimate decision. In domains like E-Elves, the team incurs such wait costs in situations where (for example) other meeting attendees have assembled in a meeting room at the time of the meeting, but are kept waiting without any input or decision from Friday (potentially because it cannot provide a high-quality decision, nor can it get any input from its user). Notice that different roles will lead to different wait cost functions, since delays in the performance of different roles will have different effects on the team. We assume that there is some point in time, $\vartriangleleft$, after which no more costs accrue, i.e., if $t \geq \vartriangleleft$ then $f(\asymp, t) = f(\asymp, \vartriangleleft)$. At the deadline, $\vartriangleleft$, the maximum cost due to inappropriate timing of a decision has been incurred. Finally, we assume that, in general, until $\vartriangleleft$, the wait cost function is non-decreasing, reflecting the idea that bigger violations of constraints lead to higher wait costs. The final element of the model is the coordination-change action, $\mathcal{D}$, which moves the agent further away from the deadline and hence reduces the wait costs that are incurred. We model the effect of the $\mathcal{D}$ by letting $\mathcal{W}$ be a function of $t - \mathcal{D}_{value}$ (rather than $t$) after the $\mathcal{D}$ action and as having a fixed cost, $\mathcal{D}_{cost}$, incurred immediately upon its execution. For example, in the E-Elves domain, suppose at the time of the meeting, Friday delays the meeting by 15 minutes ($\mathcal{D}$ action). Then, in the following time period, it will incur the relatively low cost of not making a decision 15 minutes before the meeting ($t - \mathcal{D}_{value}$), rather than the relatively high cost of not making the decision at the time of the meeting. Other, possibly more complex, models of a $\mathcal{D}$ action could also be used.

---

4. The best time to transfer control can be found, e.g., by differentiating the expected utility equation in Section 3.1 and solving for 0.

We use these four elements to compute the EU of an arbitrary strategy, $s$. The utility derived from a decision being made at time $t$ by the entity in control is the quality of the entity's decision minus the costs incurred from waiting until $t$, i.e., $EU^d_{e_c}(t) = EQ^d_{e_c}(t) - \mathcal{W}(t)$. If a coordination-change action has been taken it will also have an effect on utility. Until a coordination change of value $\mathcal{D}_{value}$ is taken at some time $\Delta$, the incurred wait cost is $\mathcal{W}(\Delta)$. Then, between $\Delta$ and $t$, the wait cost incurred is $\mathcal{W}(t - \mathcal{D}_{value}) - \mathcal{W}(\Delta - \mathcal{D}_{value})$. Thus, if a $\mathcal{D}$ action has been taken at time $\Delta$ for cost $\mathcal{D}_{cost}$ and with value $\mathcal{D}_{value}$, the utility from a decision at time $t$ ($t > \Delta$) is: $EU^d_{e_c}(t) = EQ^d_{e_c}(t) - \mathcal{W}(\Delta) - \mathcal{W}(\Delta - \mathcal{D}_{value}) + \mathcal{W}(t - \mathcal{D}_{value}) - \mathcal{D}_{cost}$. To calculate the EU of an entire strategy, we multiply the response probability mass function's value at each instant by the EU of receiving a response at that instant, and then integrate over the products. Hence, the EU for a strategy $s$ given a problem instance, $\langle A, \alpha, \rho, \asymp, d, E \rangle$, is:

$$EU_s^{\langle A, \alpha, \rho, \asymp, d, E \rangle} = \int_0^\infty P_\top(t) EU^d_{e_c}(t) \ .dt \qquad (2)$$

If a strategy involves several actions, we need to ensure that the probability of response function and the wait-cost calculation reflect the control situation at that point in the strategy. For example, if the user, $H$, has control at time $t$, $P_\top(t)$ should reflect H's

$$EU_A^d = EQ_A^d(0) - \mathcal{W}(0) \tag{3}$$

$$EU_e^d = \int_0^{\lhd} P_\top(t) \times (EQ_e^d(t) - \mathcal{W}(t)).dt + \int_{\lhd}^{\infty} P_\top(t) \times (EQ_e^d(t) - \mathcal{W}(D)).dt \tag{4}$$

$$EU_{eA}^d = \int_0^T P_\top(t) \times (EQ_e^d(t) - \mathcal{W}(t)).dt + \int_T^{\infty} P_\top(t).dt \times (EQ_a^d(T) - \mathcal{W}(T)) \tag{5}$$

$$EU_{e\mathcal{D}eA}^d = \tag{6}$$
$$\int_0^{\Delta} P_\top(t)(EQ_e^d(t) - \mathcal{W}(t)).dt' +$$
$$\int_{\Delta}^T P_\top(t)(EQ_e^d(t) - \mathcal{W}(\Delta) + \mathcal{W}(\Delta - \mathcal{D}_{value}) - \mathcal{W}(t - \mathcal{D}_{value}) - \mathcal{D}_{cost}).dt +$$
$$\int_T^{\infty} P_\top(t)(EQ_A^d(t) - \mathcal{W}(\Delta) + \mathcal{W}(\Delta - \mathcal{D}_{value}) - \mathcal{W}(T - \mathcal{D}_{value}) - \mathcal{D}_{cost}).dt$$

Table 1: General AA EU equations for sample transfer of control strategies.

probability of responding at $t$, i.e., $P_\top^H(t_0)$. To this end, we can break the integral from Equation 2 into separate terms, with each term representing one segment of the strategy, e.g., for a strategy $UA$ there would be one term for when $U$ has control and another for when $A$ has control.

Using this basic technique for writing down EU calculations, we can write down the specific equations for arbitrary transfer-of-control strategies. Equations 3-6 in Table 1 show the EU equations for the strategies $A$, $e$, $eA$ and $e\mathcal{D}eA$ respectively. The equations assume that the agent, $A$, can make the decision instantaneously (or at least, with no delay significant enough to affect the overall value of the decision). The equations are created by writing down the integral for each of the segments of the strategy, as described above. $T$ is the time when the agent takes control from $e$, and $\Delta$ is the time at which the $\mathcal{D}$ occurs. One can write down the equations for more complex strategies in the same way. Notice that these equations make no assumptions about the particular functions.

Given that the EU of a strategy can be calculated, the AA problem for the agent reduces to finding and following the transfer-of-control strategy that will maximize its EU. Formally, the agent's problem is:

**Axiom 3.1** *For a problem $\langle A, \alpha, \rho, \asymp, d, E \rangle$, the agent must select $s \in \mathbf{S}$ such that $\forall s' \in \mathbf{S}, s' \neq s, EU_s^{\langle A, \alpha, \rho, \asymp, d, E \rangle} \geq EU_{s'}^{\langle A, \alpha, \rho, \asymp, d, E \rangle}$*

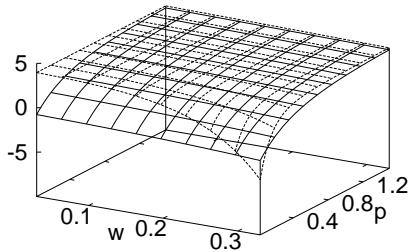

Figure 3: Graph comparing the EU of two strategies, $H\mathcal{D}A$ (solid line) and $H$ (dashed line) given a particular instantiation of the model with constant expected decision-making quality, exponentially rising wait costs, and Markovian response probabilities. $p$ is a parameter to the $P_\top(t)$ function, with higher $p$ meaning longer expected response time. $w$ is a parameter to the $\mathcal{W}(t)$ function with higher $w$ meaning more rapidly accruing wait costs.

## 3.2 Dominance Relationships among Strategies

An agent could potentially find the strategy with the highest EU by examining each and every strategy in **S**, computing its EU, and selecting the strategy with the highest value. For example, consider the problem for domains with constant expected decision-making quality, exponentially rising wait costs, and Markovian response probabilities. Figure 3 shows a graph of the EU of two strategies ($H\mathcal{D}A$ and $H$) given this particular model instantiation. Notice that, for different response probabilities and rates of wait cost accrual, one strategy outperforms the other, but neither strategy is dominant over the entire parameter space. The EU of a strategy is also dependent on the timing of transfers of control, which in turn depend on the relative quality of the entities' decision making. Appendix I provides a more detailed analysis.

Fortunately, we do not have to evaluate and compare each and every candidate in an exhaustive search to find the optimal strategy. We can instead use analytical methods to draw general conclusions about the relative values of different candidate strategies. In particular, we present three Lemmas that show the domain-level conditions under which particular strategy types are superior to others. The Lemmas also lead us to the, perhaps surprising, conclusion that complex strategies are not necessarily superior to single-shot strategies, even in a multi-agent context; in fact, no particular strategy dominates all other strategies across all domains.

Let us first consider the AA subproblem of whether an agent should *ever* take back control from another entity. If we can show that, under certain conditions, an agent should *always* eventually take back control, then our strategy selection process can ignore any strategies where the agent does not do so (i.e., any strategies not ending in $A$). The agent's goal is to strike the right balance between not waiting indefinitely for a user response and not taking a risky autonomous action. Informally, the agent reasons that it should eventually make a decision if the expected cost of continued waiting exceeds the difference between the user's decision quality and its own. More formally, the agent should eventually take back decision-making control iff, for some time $t$:

$$\int_t^\lhd P_\top(t')\mathcal{W}(t').dt' - \mathcal{W}(t) > EQ_U^d(t) - EQ_A^d(t) \tag{7}$$

where the left-hand side calculates the future expected wait costs and the right-hand side calculates the extra utility to be gained by getting a response from the user. This result leads to the following general conclusion about strategies that end with giving control back to the agent:

LEMMA 1: *If $s \in \mathbf{S}$ is a strategy ending with $e \in E$, and $s'$ is $sA$, then $EU_{s'}^d > EU_s^d$ iff $\forall e \in E, \exists t < \lhd$ such that $\int_t^\lhd P_\top(t')\mathcal{W}(t').dt' - \mathcal{W}(t) > EQ_e^d(t) - EQ_A^d(t)$*

Lemma 1 says that if, at any point in time, the expected cost of indefinitely leaving control in the hands of the user exceeds the difference in quality between the agent's and user's decisions, then strategies which ultimately give the agent control dominate those which do not. Thus, if the rate of wait cost accrual increases or the difference in the relative quality of the decision-making abilities decreases or the user's probability of response decreases, then strategies where the agent eventually takes back control will dominate. A key consequence of the Lemma (in the opposite direction) is that, if the rate that costs accrue does not accelerate, and if the probability of response stays constant (i.e., Markovian), then the agent *should* indefinitely leave control with the user (if the user had originally been given control), since the expected wait cost will not change over time. Hence, even if the agent is faced with a situation with potentially high total wait costs, the optimal strategy may be a one-shot strategy of handing over control and waiting indefinitely, because the expected future wait costs at each point in time are relatively low. Thus, Lemma 1 isolates the condition under which we should consider appending an $A$ transfer-of-control action to our strategy.

We can perform a similar analysis to identify the conditions under which we should include a $\mathcal{D}$ action in our strategy. The agent has incentive in changing coordination constraints via a $\mathcal{D}$ action due to the additional time made available for getting a high-quality response from an entity. However, the overall value of a $\mathcal{D}$ action depends on a number of factors (e.g., the cost of taking the $\mathcal{D}$ action and the timing of subsequent transfers of control). We can calculate the *expected value* of a $\mathcal{D}$ by comparing the EU of a strategy with and without a $\mathcal{D}$. The $\mathcal{D}$ is useful if and only if the increased expected value of the strategy with it is greater than its cost, $\mathcal{D}_{cost}$.

LEMMA 2: *if $s \in \mathbf{S}$ has no $\mathcal{D}$ and $s'$ is $s$ with a $\mathcal{D}$ included at $t$ then $EU_{s'}^d > EU_s^d$ iff $\int P_\top(t')\mathcal{W}(t).dt' - \int P_\top(t')\mathcal{W}(t|\mathcal{D}).dt' > \mathcal{D}_{cost}$*

We can illustrate the consequences of Lemma 2 by considering the specific problem model of Appendix I (i.e., $P_\top(t) = \rho\exp^{-\rho t}$, $\mathcal{W}(t) = \omega\exp^{\omega t}$, $EQ_e^d(t) = c$, and candidate strategies $eA$ and $e\mathcal{D}A$). In this case, $EU_{e\mathcal{D}A}^d > EU_{eA}^d$ **iff** $-(\rho - \omega)\omega\rho\exp^{-(\rho-\omega)\Delta}(1 - \exp^{-\omega\mathcal{D}_{value}}) > \mathcal{D}_{cost}$. Figure 4 plots the value of the $\mathcal{D}$ action as we vary the rate of wait cost accumulation, $w$, and the parameter of the Markovian response probability function, $p$. The graph shows that the benefit from the $\mathcal{D}$ is highest when the probability of response is neither too low nor too high. When the probability of response is low, the user is unlikely to respond, even given the extra time; hence, the agent will have incurred $\mathcal{D}_{cost}$ with no benefit. A $\mathcal{D}$ also has little value when the probability of response is high, because the user will likely respond shortly after the $\mathcal{D}$, meaning that it has little effect (the effect of the $\mathcal{D}$ is on the wait costs *after* the action is taken). Overall, according to Lemma 2, at those points where the graph goes above $\mathcal{D}_{cost}$, the agent should include a $\mathcal{D}$ action, and, at all other points, it should not. Figure 4 demonstrates the value of a $\mathcal{D}$ action for a specific subclass of problem domains, but we can extend our conclusion to the more general case as well. For instance, while the specific model has exponential wait costs, in models where wait costs grow more slowly, there will be fewer situations where Lemma 2's criterion holds (i.e., where a $\mathcal{D}$ will be useful). Thus, Lemma 2 allows us to again eliminate strategies from consideration, based on the evaluation of its criterion in the particular domain of interest.

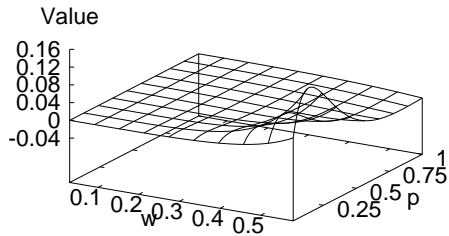

Figure 4: The value of $\mathcal{D}$ action in a particular model ($P_\top(t) = \rho \exp^{-\rho t}$, $\mathcal{W}(t) = \omega \exp^{\omega t}$, and $EQ_e^d(t) = c$).

Given Lemma 2's evaluation of adding a single $\mathcal{D}$ action to a strategy, it is natural to ask whether a second, third, etc. $\mathcal{D}$ action would increase EU even further. In other words, when a complex strategy is better than a simple one, is an even more complex strategy even better? The answer is "not necessarily".

Lemma 3: $\forall K \in \mathbf{N}, \exists \mathcal{W} \in \mathbf{W}, \exists \mathcal{P} \in \mathbf{P}, \exists \mathcal{EQ} \in \mathbf{EQ}$ *such that the optimal strategy has $K$ $\mathcal{D}$ actions.*

Informally, Lemma 3 says that we cannot fix a single, optimal number of $\mathcal{D}$ actions, because for *every* possible number of $\mathcal{D}$ actions, there is a potential domain (i.e., combination of a wait-cost, response-probability, and expected-quality functions) for which that number of $\mathcal{D}$ actions is justified by being optimal. Consider a situation where the cost of a $\mathcal{D}$ was a function of the number of $\mathcal{D}$s to date (i.e., the cost of the $K$th $\mathcal{D}$ is $f(K)$). For example, in the E-Elves' meeting case, the cost of delaying a meeting for the third time is much higher than the cost of the first delay, since each delay is successively more annoying to other meeting participants. Hence, the test for the usefulness of the $K$th $\mathcal{D}$ in a strategy, given the specific model in Appendix I, is:

$$f(K) < \omega (\exp^{-\mathcal{D}_{value}\omega} - 1) \times (\frac{\rho}{\delta}\exp^{-\delta T} - \frac{\omega}{\delta}\exp^{-\delta\Delta} - \exp^{\omega\Delta - \rho T}) \tag{8}$$

Depending on the nature of $f(K)$, Equation 8 can hold for any number of $\mathcal{D}$s, so, for any $K$, there will be some conditions for which a strategy with $K$ $\mathcal{D}$s is optimal. For instance, in Section 5.3, we show that the maximum length of the optimal strategy for a random configuration of up to 25 entities is usually less than eight actions.

Equation 8 illustrates how the value of an additional $\mathcal{D}$ can be limited by changing $\mathcal{D}_{cost}$, but Lemma 3 also shows us that other factors can affect the value of an additional $\mathcal{D}$. For example, even with a constant $\mathcal{D}_{cost}$, the value of an additional $\mathcal{D}$ depends on how many other $\mathcal{D}$ actions the agent performs. Figure 4 shows that the value of the $\mathcal{D}$ depends on the rate at which wait costs accrue. If the rate of wait cost accrual accelerates over time (e.g., for the exponential model), a $\mathcal{D}$ action slows that acceleration, rendering a second $\mathcal{D}$ action less useful (since the wait costs are now accruing more slowly). Notice also that $\mathcal{D}$s become valueless after the deadline, when wait costs stop accruing.

Taken together, Lemmas 1-3 show that no particular transfer-of-control strategy dominates all others across all domains. Moreover, very different strategies, from single-shot strategies to arbitrarily complex strategies, are appropriate for different situations, although the range of situations where a particular transfer-of-control action provides benefit can be quite narrow. Since a strategy might have very low EU for some set of parameters, choosing the wrong strategy can lead to very poor results. On the other hand, once we understand the parameter configuration of an intended application domain, Lemmas 1-3 provide useful tools for focusing the search for an optimal transfer-of-control strategy. The Lemmas can be used off-line to substantially reduce the space of strategies that need to be searched to find the optimal strategy. However, in general there may be many strategies and finding the optimal strategy may not be possible or feasible.

## 3.3 Model Predictions for the E-Elves

In this section, we use the model to predict properties of a successful approach to AA in the E-Elves. Using approximate functions for the probability of response, wait cost, and expected decision quality, we can calculate the EU of various strategies and determine the types of strategies that are going to be useful. Armed with this knowledge, we can predict some key properties of a successful implementation.

A key feature of the E-Elves is that the user is mobile. As she moves around the environment, her probability of responding to requests for decisions changes drastically, e.g., she is most likely to respond when at her workstation. To calculate the EU of different strategies, we need to know $P_{\top}(t)$, which means that we need to estimate the response probabilities and model how they change as the user moves around. When Friday communicates via a workstation dialog box, the user will respond, on average, in five minutes. However, when Friday communicates via a Palm pilot the average user response time is an hour. Users generally take longer to decide whether they want to present at a research meeting, taking approximately two days on average. So, the function $P_{\top}(t)$ should have an average value of 5 minutes when the user in her office, an average of one hour when the user is contacted via a Palm pilot and an average of two days when the decision is whether to present at a research meeting. It is also necessary to estimate the relative quality of the user, $EQ_U^d(t)$, and Friday's decision making, $EQ_A^d(t)$. We assume that the user's decision-making $EQ_U^d(t)$ is high with respect to Friday's, $EQ_A^d(t)$. The uncertainty about user intentions makes it very hard for Friday to consistently make correct decisions about the time at which the user will arrive at meetings, although its sensors (e.g., GPS device) give some indication of the user's location. When dealing with more important meetings, the cost of Friday's errors is higher. Thus, in some cases, the decision-making quality of the user and Friday will be similar, i.e., $EQ_d^U(t) \approx EQ_d^A(t)$; while in other cases, there will be an order of magnitude difference, i.e., $EQ_d^U(t) \approx 10 * EQ_d^A(t)$. The wait cost function, $\mathcal{W}(t)$, will be much larger for big meetings than small and increase rapidly as other attendees wait longer in the meeting room. Finally, the cost of delays, i.e., $\mathcal{D}_{cost}$, can vary by about an order of magnitude. In particular, the cost of rescheduling meetings varies greatly, e.g., the cost of rescheduling small informal meetings with colleagues is far less than rescheduling a full lecture room at 5 PM Friday.

The parameters laid out above show how parameters vary from decision to decision. For a specific decision, we use Markovian response probabilities (e.g., when the user is in her office, the average response time is five minutes), exponentially increasing wait costs, and constant decision-making quality (though it changes from decision to decision) to calculate the EU of interesting strategies. Calculating the EU of different strategies using the values for different parameters shown above allows us to draw the following conclusions (Table 5 in Section 5.3 presents a quantitative illustration of these predictions):

- The strategy $e$ should not be used, since for all combinations of user location and meeting importance the EU of this strategy is very low.

- Multiple strategies are required, since for different user locations and meeting importance different strategies are optimal.

- Since quite different strategies are required when the user is in different locations, the AA reasoning will need to change strategies when the user changes location.

- No strategy has a reasonable EU for all possible parameter instantiations, hence always using the same strategy will occasionally cause dramatic failures.

- For most decisions, strategies will end with the agent taking a decision, since strategies ending with the user in control generally have very low EU.

These predictions provide important guidance about a successful solution for AA in the E-Elves. In particular, they make clear that the approach must flexibly choose between different strategies and adjust depending on the meeting type and user location.

Section 2.2 described the unsuccessful C4.5 approach to AA in E-Elves and identified several reasons for the mistakes that occurred. In particular, rigidly transferring control to one entity and ignoring potential team costs involved in an agent's decision were highlighted as reasons for the dramatic mistakes in Friday's autonomy reasoning. Reviewing the C4.5 approach in the light of the notion of strategies, we see that Friday learned one strategy and stuck with that strategy. In particular, originally, Friday would wait indefinitely for a user response, i.e., it would follow strategy $e$, if it had learned to transfer control. As shown later in Table 5, this strategy has a very low EU. When a fixed-length timeout was introduced, Friday would follow strategy $e(5)A$. Such a strategy has high EU when $EQ_d^U(t) \approx EQ_d^A(t)$ but very low EU when $EQ_d^U(t) \approx 10 * EQ_d^A(t)$. Thus, the model explains a phenomenon observed in practice.

On the other hand, we can use the model to understand that C4.5's failure in this case does not mean that it will never be useful for AA. Different strategies are only required when certain parameters (like probability of response or wait cost) change significantly. In applications where such parameters do not change dramatically from decision to decision, one particular strategy may always be appropriate. For such applications, C4.5 might learn the right strategy just with a small amount of training data and perform acceptably well.

## 4. Operationalizing Strategies with MDPs

We have formalized the problem of AA as the selection of the transfer-of-control strategy with the highest EU. We now need an operational mechanism that allows an agent to perform that selection. One major conclusion from the previous section is that different strategies dominate in different situations, and that applications such as E-Elves will require mechanism(s) for selecting strategies in a situation-sensitive fashion. In particular, the mechanism must flexibly change strategies as the situation changes. The required mechanism must also represent the utility function specified by our expected decision qualities, **EQ**, the costs of violating coordination constraints, **W**, and our coordination-change cost, $\mathcal{D}_{cost}$. Finally, the mechanism must also represent the uncertainty of entity responses and then look ahead over the possible responses (or lack thereof) that may occur in the future.

MDPs are a natural means of performing the decision-theoretic planning required to find the best transfer-of-control strategy. MDP policies provide a mapping between the agent's state and the optimal transfer of control strategy. By encoding the parameters of the model of AA strategies into the MDP, the MDP effectively becomes a detailed implementation of the model and, hence, assumes its properties. We can use standard algorithms (Puterman, 1994) to find the optimal MDP policy and, hence, the optimal strategies to follow in each state.

To simplify exposition, as well as to illustrate the generality of the resulting MDP, this section describes the mapping from AA strategies to the MDP in four subsections. In particular, Section 4.1 provides a direct mapping of strategies to an abstract MDP. Section 4.2 fills in state features to enable a more concrete realization of the reward function, while still maintaining a domain-independent view. Thus, the section completely defines a general MDP for AA is potentially reusable across a broad class of domains. Section 4.3 illustrates an implemented instantiation of the MDP in E-Elves. Section 4.4 addresses further practical issues in operationalizing such MDPs in domains such as E-Elves.

### 4.1 Abstract MDP Representation of AA Problem

Our MDP representation's fundamental state features capture the state of control:

- *controlling-entity* is the entity that currently has decision-making control.

- *$e_i$-response* is any response $e_i$ has made to the agent's requests for input.

| Original State | | Action | Destination State | | | Probability |
|---|---|---|---|---|---|---|
| $e_{ctrl}$ | $time$ | | $e_{ctrl}$ | $e_i$-response | $time$ | |
| $e_j$ | $t_k$ | $e_i$ | $e_i$ | yes | $t_{k+1}$ | $\int_{t_k}^{t_{k+1}} P_\top^{e_i}(t)dt$ |
| $e_j$ | $t_k$ | $e_i$ | $e_i$ | no | $t_{k+1}$ | $1 - \int_{t_k}^{t_{k+1}} P_\top^{e_i}(t)dt$ |
| $e_i$ | $t_k$ | wait | $e_i$ | yes | $t_{k+1}$ | $\int_{t_k}^{t_{k+1}} P_\top^{e_i}(t)dt$ |
| $e_i$ | $t_k$ | wait | $e_i$ | no | $t_{k+1}$ | $1 - \int_{t_k}^{t_{k+1}} P_\top^{e_i}(t)dt$ |
| $e_i$ | $t_k$ | $\mathcal{D}$ | $e_i$ | no | $t_k - \mathcal{D}_{value}$ | $1$ |

Table 2: Transition probability function for AA MDP. $e_{ctrl}$ is the *controlling-entity*.

- *time* is the current time, typically discretized and ranging from 0 to our deadline, $\triangleleft$ — i.e., a set $\{t_0 = 0, t_1, t_2, \ldots, t_n = \triangleleft\}$.

If $e_i$-*response* is not null or if *time* $= \triangleleft$, then the agent is in a terminal state. In the former case, the decision is the value of $e_i$-*response*.

We can specify the set of actions for this MDP representation as $\Gamma = E \cup \{\mathcal{D}, \text{wait}\}$. The set of actions subsumes the set of entities, $E$, since the agent can transfer decision-making control to any one of these entities. The $\mathcal{D}$ action is the coordination-change action that changes coordination constraints, as discussed earlier. The "wait" action puts off transferring control and making any autonomous decision, without changing coordination with the team. The agent should reason that "wait" is the best action when, in time, the situation is likely to change to put the agent in a position for an improved autonomous decision or transfer-of-control, without significant harm. For example, in the E-Elves domain, at times closer to a meeting, users can generally make more accurate determinations about whether they will arrive on time, hence it is sometimes useful to wait when the meeting is a long time off.

The transition probabilities (specified in Table 2) represent the effects of the actions as a distribution over their effects (i.e., the ensuing state of the world). If, in a state with *time* $= t_k$, the agent chooses an action that transfers decision-making control to an entity, $e_i$, other than the agent itself, the outcome is a state with *controlling-entity* $= e_i$ and *time* $= t_{k+1}$. There are two possible outcomes for $e_i$-*response*: either the entity responds with a decision during this transition (producing a terminal state), or it does not, and we derive the probability distribution over the two from **P**. The "wait" action has a similar branch, except that the *controlling-entity* remains unchanged. Finally, the $\mathcal{D}$ action occurs instantaneously, so there is no time for the controlling entity to respond, but the resulting state effectively moves to an earlier time (e.g., from $t_k$ to $t_k - \mathcal{D}_{value}$).

We can derive the reward function for this MDP in a straightforward fashion from our strategy model. Table 3 presents the complete specification of this reward function. In transitions that take up time, i.e., transferring control and not receiving a response (Table 3, row 1) or "wait" (Table 3, row 2), the agent incurs the wait cost of that interval. In transitions where the agent performs $\mathcal{D}$, the agent incurs the cost of that action (Table 3, row 3). In terminal states with a response from $e_i$, the agent derives the expected quality of that entity's decision (Table 3, row 4). A policy that maximizes the reward that an agent expects to receive according to this AA MDP model will correspond exactly to an optimal transfer-of-control strategy. Note that this reward function is described in an abstract fashion—for example, it does not specify how to compute the agent's expected quality of decision, $EQ_d^A(t)$.

| controlling-entity | time | $e_i$-response | Action | Reward |
|:---:|:---:|:---:|:---:|:---:|
| $e_j$ | $t_k$ | no | $e_i$ | $\mathcal{W}(k+1) - \mathcal{W}(k)$ |
| $e_i$ | $t_k$ | no | wait | $\mathcal{W}(k+1) - \mathcal{W}(k)$ |
| $e_i$ | $t_k$ | no | $\mathcal{D}$ | $\mathcal{D}_{cost}$ |
| $e_i$ | $t_k$ | yes | | $EQ_{e_i}^d(t_k)$ |

Table 3: Reward function for AA MDP.

## 4.2 MDP Representation of AA Problem within Team Context

We have now given a high-level description of an MDP for implementing the notion of transfer-of-control strategies for AA. The remainder of this section provides a more detailed look at the MDP for a broad class of AA domains (including the E-Elves) where the agent acts on behalf of a user who is filling a role, $\rho$, within the context of a team activity, $\alpha$. The reward function compares the EU of different strategies, finding the optimal one for the current state. To facilitate this calculation, we need to represent the parameters used in the model. We introduce the following state features to capture the aspects of the AA problem in a team context:

- *team-orig-expect-$\rho$* is what the team originally expected of the fulfilling of $\rho$.

- *team-expect-$\rho$* is the team's current expectations of what fulfilling the role $\rho$ implies.

- *agent-expect-$\rho$* is the agent's (probabilistic) estimation for how $\rho$ will be fulfilled.

- "other $\alpha$ attributes" encapsulate other aspects of the joint activity that are affected by the decision.

When we add these more specific features to the generic AA state features already presented, the overall state, within the MDP representation of a decision $d$, is a tuple:

$$\langle controlling\text{-}entity, team\text{-}orig\text{-}expect\text{-}\rho, team\text{-}expect\text{-}\rho, agent\text{-}expect\text{-}\rho, \alpha\text{-}status,$$
$$e_i\text{-}response, time, \text{other } \alpha \text{ attributes}\rangle$$

For example, for a meeting scenario, *team-orig-expect-$\rho$* could be "Meet at 3pm", *team-expect-$\rho$* could be "Meet at 3:15pm" after a user requested a delay, and *agent-expect-$\rho$* could be "Meet at 3:30pm" if the agent believes its user will not make the rescheduled meeting.

The transition probability function for the AA MDP in a team context includes our underlying AA transition probabilities from Table 3, but it must also include probabilities over these new state features. In particular, in addition to the temporal effect of the $\mathcal{D}$ action described in Section 4.1, there is the additional effect on the coordination of $\alpha$. The $\mathcal{D}$ action changes the value of the *team-expect-$\rho$* feature (in a domain-dependent but deterministic way). No other actions affect the team's expectations. The *team-orig-expect-ρ* feature does not change; we include it to simplify the definition of the reward function. The transition probabilities over *agent-expect-ρ* and other $\alpha$-specific features are domain-specific. We provide an example of such transition probabilities in Section 4.3.

The final part of the MDP representation is the reward function. Our team AA MDP framework uses a reward function that breaks down the function from Table 3 as follows:

$$
\begin{aligned}
R(s,a) \; = \;\; & f(\textit{team-orig-expect-}\rho(s), \textit{team-expect-}\rho(s), \textit{agent-expect-}\rho(s), \\
& \alpha\textit{-status}(s), \textit{time}(s), a) \hspace{3cm} (9) \\
= \;\; & \sum_{e \in E \setminus \{A\}} EQ_e^d(\textit{time}(s)) \cdot e\textit{-response} \\
& - \lambda_1 f_1(\| \textit{ team-orig-expect-}\rho(s) - \textit{team-expect-}\rho(s) \ \|) \\
& - \lambda_{21} f_{21}(\textit{time}(s)) \\
& - \lambda_{22} f_{22}(\| \textit{ team-expect-}\rho(s) - \textit{agent-expect-}\rho(s) \ \|) \\
& + \lambda_3 f_3(\alpha\textit{-status}(s)) + \lambda_4 f_4(a)
\end{aligned}
$$

$$(10)$$

The first component of the reward function captures the value of getting a response from a decision-making entity other than the agent itself. Notice that only one entity will actually respond, so only one *e-response* will be non-zero. This corresponds to the $EQ_d^e(t)$ function used in the model and the bottom row of Table 3. The $f_1$ function reflects the inherent value of performing a role as the team originally expected, hence deterring the agent from taking costly coordination changes unless they can gain some indirect value from doing so. This corresponds to $\mathcal{D}_{cost}$ from the mathematical model and the third row of Table 3. The $f_{21}$ corresponds to the second row of Table 3, so it represents the wait cost function, $\mathcal{W}(t)$, from the model. This component encourages the agent to keep other team members informed of the role's status (e.g., by making a decision or taking an explicit $\mathcal{D}$ action), rather than causing them to wait without information. Functions $f_{22}$ and $f_3$ represent the quality of the agent's decision, represented by $Q_d^A(t)$. The standard MDP algorithms compute an expectation over the agent's reward, and an expectation over this quality will produce the desired $EQ_d^A(t)$ from the fourth row of Table 3. The first quality function, $f_{22}$, reflects the value of keeping the team's understanding of how the role will be performed in accordance with how the agent expects the user to actually perform the role. The agent receives most reward when the role is performed exactly as the team expects, but because of the uncertainty in the agent's expectation, errors are possible. $f_{22}$ represents the costs that come with such errors. The second quality component, $f_3$, influences overall reward based on the successful completion of the joint activity, which encourages the agent to take actions that maximize the likelihood that the joint activity succeeds. The desire to have the joint task succeed is implicit in the mathematical model but must be explicitly represented in the MDP. The component, $f_4$, augments the first row from Table 3 to account for additional costs of transfer-of-control actions. In particular, $f_4$ can be broken down further as follows:

$$
f_4(a) = \begin{cases} q(e) & \text{if } a \in E \\ 0 & \text{otherwise} \end{cases} \hspace{3cm} (11)
$$

The function $q(e)$ represents the cost of transferring control to a particular entity, e.g., the cost of a WAP phone message to a user. Notice, that these detailed, domain-specific costs do not appear directly in the model.

Given the MDP's state space, actions, transition probabilities, and reward function, an agent can use *value iteration* to generate a policy $P : S \rightarrow \Gamma$ that specifies the optimal action in each state (Puterman, 1994). The agent then executes the policy by taking the action that the policy dictates in each and every state in which it finds itself. A policy may include several transfers of control and coordination-change actions. The particular series of actions depends on the activities of the user. We can then interpret this policy as a contingent combination of many transfer-of-control strategies, with the strategy to follow chosen depending on the user's status (i.e., *agent-expect-ρ*).

### 4.3 Example: The E-Elves MDPs

An example of an AA MDP is the generic *delay MDP*, which can be instantiated for any meeting for which Friday may act on behalf of its user. Recall the decision, $d$, is whether to let other meeting attendees wait for a user or to begin their meeting. The joint activity, $\alpha$, is the meeting in which the agent has the role, $\rho$, of ensuring that its user attends the meeting at the scheduled time. The coordination constraints, $\asymp$, are that the attendees arrive at the meeting location simultaneously and the effect of the $\mathcal{D}$ action is to delay or cancel the meeting.

In the delay MDP's state representation, *team-orig-expect-ρ* is *originally-scheduled-meeting-time*, since attendance at the originally scheduled meeting time is what the team originally expects of the user and is the best possible outcome. *team-expect-ρ* is *time-relative-to-meeting*, which may increase if the meeting is delayed. *α-status* becomes *status-of-meeting*. *agent-expect-ρ* is not represented explicitly; instead, *user-location* is used as an observable heuristic of when the user is likely to attend the meeting. For example, a user who is away from the department shortly before a meeting should begin is unlikely to be attending on time, if at all. With all the state features, the total state space contains 2800 states for each individual meeting, with the large number of states arising from a very fine-grained discretization of time.

The general reward function is mapped to the *delay MDP* reward function in the following way.

$$f_1 = \begin{cases} g(N, \alpha) & \text{if } N < 4 \\ \infty & \text{otherwise} \end{cases} \tag{12}$$

where $N$ is the number of times the meeting is rescheduled and $g$ is a function that takes into account factors like the number of meeting attendees, the size of the meeting delay and the time until the originally scheduled meeting time. This function effectively forbids the agent from ever performing 4 or more $\mathcal{D}$ actions.

In the *delay MDP*, the functions, $f_{21}$ and $f_{22}$, both correspond to the cost of making the meeting attendees wait, so we can merge them into a single function, $f_2$. We expect that such a consolidation is possible in similar domains where the team's expectations relate to the temporal aspect of role performance.

$$f_2 = \begin{cases} h(late, \alpha) & \text{if } late > 0 \\ 0 & \text{otherwise} \end{cases} \quad (13)$$

where *late* is the difference between the scheduled meeting time and the time the user arrives at the meeting room. *late* is probabilistically calculated by the MDP based on the user's current location and a model of the user's behavior.

$$f_3 = \begin{cases} r_\alpha + r_{user} & \text{if the user attends} \\ r_\alpha & \text{if the meeting takes place, but the user does not attend} \\ 0 & \text{otherwise} \end{cases} \quad (14)$$

The value, $r_\alpha$, models the inherent value of $\alpha$, while the value $r_{user}$ models the user's individual value to $\alpha$.

$f_4$ was given previously in Equation 11. The cost of communicating with the user depends on the medium which is used to communicate. For example, there is higher cost to communicating via a WAP phone than via a workstation dialog box.

When the users are asked for input, it is assumed that, if they respond, their response will be "correct", i.e., if a user says to delay the meeting by 15 minutes, we assume the user will arrive on time for the re-scheduled meeting. If the user is asked while in front of his/her workstation, a dialog like the one shown in Figure 5 is popped up, allowing the user to select the action to be taken. The expected quality of the agent's decision is calculated by considering the agent's proposed decision and the possible outcomes of that decision. For example, if the agent proposes delaying the meeting by 15 minutes, the calculation of the decision quality includes the probability and benefits that the user will actually arrive 15 minutes after the originally scheduled meeting time, the probability and costs that the user arrives at the originally scheduled meeting time, etc.

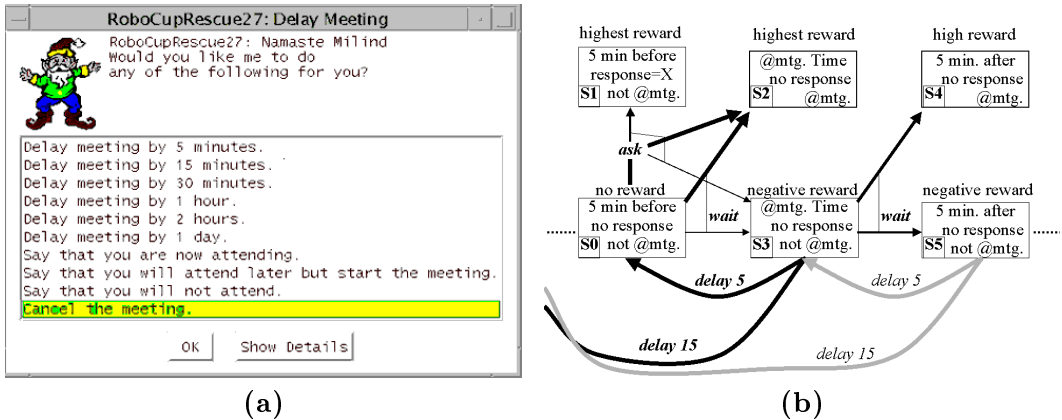

**(a)**　　　　　　　　　　　　　　**(b)**

Figure 5: (a) Dialog box for delaying meetings. (b) A small portion of the *delay MDP*.

The delay MDP also represents probabilities that a change in user location (e.g., from office to meeting location) will occur in a given time interval. Figure 5(b) shows a portion of the state space, showing only the *user-response*, and *user location* features. A transition labeled "delay $n$" corresponds to the action "delay by $n$ minutes". The figure also shows multiple transitions due to "ask" (i.e., transfer control to the user) and "wait" actions, where the relative probability of each outcome is represented by the thickness of the arrow. Other state transitions correspond to uncertainty associated with a user's response (e.g., when the agent performs the "ask" action, the user may respond with specific information or may not respond at all, leaving the agent to effectively "wait"). One possible policy produced by the delay MDP, for a subclass of meetings, specifies "ask" in state **S0** of Figure 5(b) (i.e., the agent gives up some autonomy). If the world reaches state **S3**, the policy specifies "wait". However, if the agent then reaches state **S5**, the policy chooses "delay 15", which the agent then executes autonomously. In terms of strategies, this sequence of actions is $H\mathcal{D}$.

Earlier, we described another AA decision in the E-Elves, namely whether to close an auction for an open team role. Here, we briefly describe the key aspects of the mapping of that decision to the MDP. The auction must be closed in time for the user to prepare for the meeting, but with sufficient time given for interested users to submit bids and for the human team leader to choose a particular user. *team-orig-expect-*$\rho(s)$ is that a high-quality presenter be selected with enough time to prepare. There is no $\mathcal{D}$ action, hence *team-expect-*$\rho(s) = $ *team-orig-expect-*$\rho(s)$. *agent-expect-*$\rho(s)$ is whether the agent believes it has a high-quality bid or believes such a bid will arrive in time for that user to be allocated to the role. The agent's decision quality, $EQ_A^d(t)$, is a function of the number of bids that have been submitted and the quality of those bids, e.g., if all team members have submitted bids and one user's bid stands out, the agent can confidently choose that user to do the presentation. Thus, $\alpha$-status is primarily the quality of the best bid so far and the difference between the quality of that bid and the second-best bid. The most critical component of the reward function from Equation 10 is the $\lambda_2$ component, which gives reward if the agent fulfills the users' expectation of having a willing presenter do a high-quality presentation.

## 4.4 User-Specified Constraints

The standard MDP algorithms provide the agent with optimal policies subject to the encoded probabilities and reward function. Thus, if the agent designer has access to correct models of the entities' (e.g., human users in the E-Elves) decision qualities and probabilities of response, then the agent will select the best possible transfer-of-control strategy. However, it is possible that the entities themselves have more accurate information about their own abilities than does the agent designer. To exploit this knowledge, an entity could communicate its model of its quality of decision and probability of response directly to the agent designer. Unfortunately, the typical entity is unlikely to be able to express its knowledge in the form of our MDP reward function and transition probabilities. An agent could potentially learn this additional knowledge on its own through its interactions with the entities in the domain. However, learning may require an arbitrarily large number of such interactions, all of which will take place without the benefit of the entities' inside knowledge.

As an alternative, we can provide a language of *constraints* that allows the entities to directly and immediately communicate their inside information to the agent. Our constraint language provides the entities a simple way to inform the agent of their specific properties and needs. An entity can use a constraint to forbid the agent from entering specific states or performing specific actions in specific states. Such constraints can be directly communicated by a user via the tool shown in Figure 6. For instance, in the figure shown the user is forbidding the agent from autonomous action five minutes before the meeting. We define such **forbidden-action** constraints to be a set, $C_{fa}$, where each element constraint is a boolean function, $c_{fa} : S \times A \to \{t, f\}$. Similarly, we define **forbidden-state** constraints to be a set, $C_{fs}$, with elements, $c_{fs} : S \to \{t, f\}$. If a constraint returns $t$ for a particular domain element (either state or state-action pair, as appropriate), then the constraint applies to the given element. For example, a forbidden-action constraint, $c_{fa}$, forbids the action $a$ from being performed in state $s$ if and only if $c_{fa}(s, a) = t$.

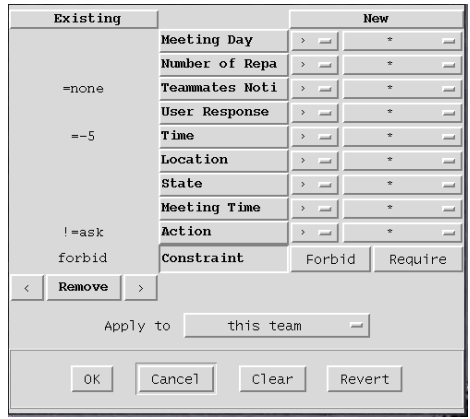

Figure 6: Screenshot of the tool for entering constraints. The constraint displayed forbids not transferring control (i.e., forces transfer) five minutes before the meeting if the teammates have previously been given information about the user's attendance at the meeting.

To provide probabilistic semantics, suitable for an MDP context, we first provide some notation. Denote the probability that the agent will ever arrive in state $s_f$ after following a policy, $P$, from an initial state $s_i$ as $\Pr(s_i \xrightarrow{*} s_f | P)$. Then, we define the semantics of a forbidden-state constraint $c_{fs}$ as requiring $\Pr(s_i \xrightarrow{*} s_f | P) = 0$. The semantics given to a forbidden-action constraint, $c_{fa}$, is a bit more complex, requiring $\Pr(s_i \xrightarrow{*} s_f \wedge P(s_f)=a | P) = 0$ (i.e., $c_{fa}$ forbids the agent from entering state $s_f$ *and* then performing action $a$). In some cases, an aggregation of constraints may forbid *all* actions in state $s_f$. In this case, the conjunction allows the agent to still satisfy all forbidden-action constraints by avoiding $s_f$ (i.e., the state $s_f$ itself becomes forbidden). Once a state, $s_f$, becomes indirectly forbidden in this fashion, any action that potentially leads the agent from an ancestor state into $s_f$ likewise becomes forbidden. Hence, the effect of forbidding constraints can propagate backward through the state space, affecting state/action pairs beyond those which cause immediate violations.

The forbidding constraints are powerful enough for the entity to communicate a wide range of knowledge about their decision quality and probability of response to the agent. For instance, some E-Elves users have forbidden their agents from rescheduling meetings to lunch time. To do so, the users provide a feature specification of the states they want to forbid, such as *meeting-time*=12 PM. Such a specification generates a forbidden-state constraint, $c_{fs}$, that is true in any state, $s$, where *meeting-time*=12 PM in $s$. This constraint effectively forbids the agent from performing any $\mathcal{D}$ action that would result in a state where *meeting-time*=12PM. Similarly, some users have forbidden autonomous actions in certain states by providing a specification of the actions they want to forbid, e.g., *action*≠"ask". This generates a forbidden-action constraint, $c_{fa}$, that is true for any state/action pair, $(s, a)$, with $a \neq$"ask". For example, a user might specify such a constraint for states where they are in their office, at the time of a meeting because they know that they will always make decisions in that case. Users can easily create more complicated constraints by specifying values for multiple features, as well as by using comparison functions other than = (e.g., ≠, >).

Analogous to the forbidding constraints, we also introduce **required-state** and **required-action** constraints, defined as sets, $C_{rs}$ and $C_{ra}$, respectively. The interpretation provided to the required-state constraint is symmetric, but opposite to that of the forbidden-state constraint: $\Pr(s_i \overset{*}{\to} s_f | P) = 1$. Thus, from any state, the agent *must* eventually reach a required state, $s_f$. Similarly, for the required-action constraint, $\Pr(s_i \overset{*}{\to} s_f \wedge P(s_f) = a | P) = 1$. The users specify such constraints as they do for their forbidding counterparts (i.e., by specifying the values of the relevant state features or action, as appropriate). In addition, the requiring constraints also propagate backward. Informally, the forbidden constraints focus locally on specific states or actions, while the required constraints express global properties over all states.

The resulting language allows the agent to exploit synergistic interactions between its initial model of transfer-of-control strategies and entity-specified constraints. For example, a forbidden-action constraint that prevents the agent from taking autonomous action in a particular state is equivalent to the user specifying that the agent must transfer control to the user in that state. In AA terms, the user instructs the agent not to consider any transfer-of-control strategies that violate this constraint. To exploit this pruning of the strategy space by the user, we have extended standard value iteration to also consider constraint satisfaction when generating optimal strategies. Appendix II provides a description of a novel algorithm that finds optimal policies while respecting user constraints. The appendix also includes a proof of the algorithm's correctness.

## 5. Experimental Results

This section presents experimental results aimed at validating the claims made in the previous sections. In particular, the experiments aim to show the utility of complex transfer-of-control strategies and the effectiveness of MDPs as a technique for their operationalization. Section 5.1 details the use of the E-Elves in daily activities and Section 5.2 discusses the pros and cons of living and working with the assistance of Fridays. Section 5.3 shows some characteristics of strategies in this type of domain (in particular, that different strategies are used in practice). Finally, Section 5.4 describes detailed experiments that illustrate characteristics of the AA MDP.

## 5.1 The E-Elves in Daily Use

The E-Elves system was heavily used by ten users in a research group at ISI, between June 2000 and December 2000.[5] The Friday agents ran continuously, around the clock, seven days a week. The exact number of agents running varied over the period of execution, with usually five to ten Friday agents for individual users, a capability matcher (with proxy), and an interest matcher (with proxy). Occasionally, temporary Friday agents operated on behalf of special guests or other short-term visitors.

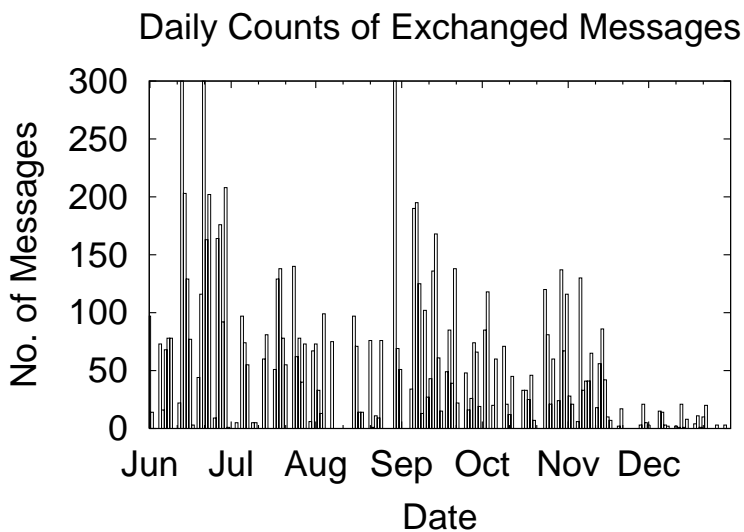

Figure 7: Number of daily coordination messages exchanged by proxies over a seven-month period.

Figure 7 plots the number of daily messages exchanged by the Fridays over seven months (June through December, 2000). The size of the daily counts reflects the large amount of coordination necessary to manage various activities, while the high variability illustrates the dynamic nature of the domain (note the low periods during vacations and final exams). Figure 8(a) illustrates the number of meetings monitored for each user. Over the seven months, nearly 700 meetings where monitored. Some users had fewer than 20 meetings, while others had over 250. Most users had about 50% of their meetings delayed (this includes regularly scheduled meetings that were cancelled, for instance due to travel). Figure 8(b) shows that usually 50% or more of delayed meetings were autonomously delayed. In this graph, repeated delays of a single meeting are counted only once. The graphs show that the

5. The user base of the system was greatly reduced after this period due to personnel relocations and student graduations, but it remains in use with a smaller number of users.

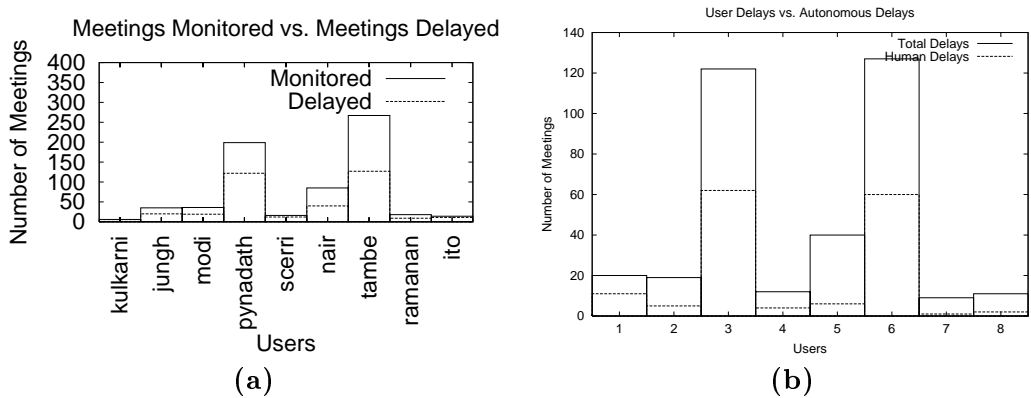

Figure 8: (a) Monitored vs. delayed meetings per user. (b) Meetings delayed autonomously vs. by hand.

agents are acting autonomously in a large number of instances, but, equally importantly, humans are also often intervening, indicating the critical importance of *adjustable* autonomy in Friday agents.

For a seven-month period, the presenter for USC/ISI's TEAMCORE research group presentations was decided using auctions. Table 4 shows a summary of the auction results. Column 1 ("Date") shows the dates of the research presentations. Column 2 ("No. of Bids") shows the total number of bids received before a decision. A key feature is that auction decisions were made without all 9 users entering bids; in fact, in one case, only 4 bids were received. Column 3 ("Best bid") shows the winning bid. A winner typically bid < 1, 1 >, i.e., indicating that the user it represents is both capable and willing to do the presentation — a high-quality bid. Interestingly, the winner on July 27 made a bid of < 0, 1 >, i.e., not capable but willing. The team was able to settle on a winner despite the bid not being the highest possible, illustrating its flexibility. Finally, columns 4 ("Winner") and 5 ("Method") show the auction outcome. An 'H' in column 5 indicates the auction was decided by a human, an 'A' indicates it was decided autonomously. In five of the seven auctions, a user was automatically selected to be presenter. The two manual assignments were due to exceptional circumstances in the group (e.g., a first-time visitor), again illustrating the need for AA.

| Date | No. of bids | Best bid | Winner | Method |
|------|-------------|----------|--------|--------|
| Jul 6, 2001 | 7 | 1,1 | Scerri | H |
| Jul 20, 2001 | 9 | 1,1 | Scerri | A |
| Jul 27, 2001 | 7 | 0,1 | Kulkarni | A |
| Aug 3, 2001 | 8 | 1,1 | Nair | A |
| Aug 3, 2001 | 4 | 1,1 | Tambe | A |
| Sept 19, 2001 | 6 | -,- | Visitor | H |
| Oct 31, 2001 | 7 | 1,1 | Tambe | A |

Table 4: Results for auctioning research presentation slot.

## 5.2 Evaluating the Pros and Cons of E-Elves Use

The general effectiveness of the E-Elves is shown by several observations. During the E-Elves' operation, the group members exchanged very few email messages to announce meeting delays. Instead, Fridays autonomously informed users of delays, thus reducing the overhead of waiting for delayed members. Second, the overhead of sending emails to recruit and announce a presenter for research meetings was assumed by agent-run auctions. Third, a web page, where Friday agents post their users' location, was commonly used to avoid the overhead of trying to track users down manually. Fourth, mobile devices kept users informed remotely of changes in their schedules, while also enabling them to remotely delay meetings, volunteer for presentations, order meals, etc. Users began relying on Friday so heavily to order lunch that one local "Subway" restaurant owner even suggested: "... *more and more computers are getting to order food... so we might have to think about marketing to them!!*". Notice that this daily use of the E-Elves by a number of different users occurred only after the MDP implementation of AA replaced the unreliable C4.5 implementation.

However, while the agents ensured that users spent less time on daily coordination (and miscoordination), there was a price to be paid. One issue was that users felt they had less privacy when their location was continually posted on the web and monitored by their agent. Another issue was the security of private information such as credit card numbers used for ordering lunch. As users adjusted to having agents monitor their daily activities, some users adjusted their own behavior around that of the agent. One example of such behavior was some users preferring to be a minute or two early for a meeting lest their agent decide they were late and delay the meeting. In general, since the agents never made catastrophically bad decisions most users felt comfortable using their agent and frequently took advantage of its services.

The most emphatic evidence of the success of the MDP approach is that, since replacing the C4.5 implementation, the agents have *never* repeated any of the catastrophic mistakes enumerated in Section 2.2. In particular, Friday avoids errors such as error 3 from Section 2.2 by selecting a strategy with a single, large $\mathcal{D}$ action, because it has a higher EU than a strategy with many small $\mathcal{D}$s (e.g., $\mathcal{DDDD}$). Friday avoids error 1, because the large cost associated with an erroneous cancel action significantly penalizes the EU of a cancellation. Friday instead chooses the higher-EU strategy that first transfers control to a user before taking such an action autonomously. Friday avoids errors such as errors 2 and 4 by selecting strategies in a situation-sensitive manner. For instance, if the agent's decision-making quality is low (i.e., high risk), then the agent can perform a coordination-change action to allow more time for user response or for the agent itself to get more information. In other words, it flexibly uses strategies like $e\mathcal{D}eA$, rather than always using the $e(5)A$ strategy discussed in Section 2.2. This indicates that a reasonably appropriate strategy was chosen in each situation. Although the current agents do occasionally make mistakes, these errors are typically on the order of transferring control to the user a few minutes earlier than may be necessary. Thus, the agents' decisions have been reasonable, though not always optimal.[6]

---

6. The inherent subjectivity in user feedback makes a determination of optimality difficult.

### 5.3 Strategy Evaluation

The previous section looked at the application of the MDP approach to the E-Elves but did not address strategies in particular. In this section, we specifically examine strategies in the E-Elves. We show that Fridays did indeed follow strategies and that the strategies followed were the ones predicted by the model. We also show how the model led to an insight that, in turn, led to a dramatic simplification in one part of the implementation. Finally, we show that the use of strategies is not limited to the E-Elves application by showing empirically that, for random configurations of entities, the optimal strategy will have more than one transfer-of-control action in 70% of cases.

Figure 9 shows a frequency distribution of the number of actions taken per meeting (this graph omits "wait" actions). The number of actions taken for a meeting corresponds to the length of the part of the strategy followed (the strategy may have been longer, but a decision was made so the actions were not taken). The graph shows both that the MDP followed complex strategies in the real world and that it followed *different* strategies at different times. The graph bears out the model's predictions that different strategies would be required of a good solution to the AA problem in the E-Elves domain.

Table 5 shows the EU values computed by the model and the strategy selected by the MDP. Recall that the MDP explicitly models the users' movements between locations, while the model assumes that the users do not move. Hence, in order to do an accurate comparison between the model and the MDP's results, we focus on only those cases when the user's location does not change (i.e., where the probability of response is constant). These EU values were calculated using the parameter values set out in Section 3.3. Notice, that the MDP will often perform $\mathcal{D}$s before transferring control to buy time to reduce uncertainty. The model is an abstraction of the domain, so such $\mathcal{D}$ actions, like changes in user location, are not captured. Except for a slight discrepancy in the first case the match between the MDP's behavior and the model's predictions is exact, provided that we ignore the $\mathcal{D}$ actions at the beginning of some MDP strategies. Thus, despite the model being considerably abstracted from the domain there is high correlation between the MDP policies and the model's suggested strategies. Moreover, general properties of the policies that were predicted by the model were borne out exactly. In particular, recall that the model predicted different strategies would be required, that strategy $e$ would not be used, and that generally strategies ending in $A$ would be best — all properties of the MDP policies.

The model predicts that if parameters do not vary greatly then it is sufficient to find a single optimal strategy and follow that strategy in each situation. The MDP for the decision to close an auction is an instance of this for the E-Elves. The same pattern of behavior is followed every time an open role needs to be filled by the team. This consistency arises because the wait cost is the same (since the meetings are the same) and because the pattern of incoming bids is reasonably consistent (variations in individuals' behavior cancel each other out when we look at the team as a whole). The model predicts that when parameters do not change, we can find the optimal strategy for those parameters and execute that strategy every time. However, since the MDP had worked effectively for the meeting AA, an MDP was also chosen for implementing the auction AA. When it was realized that the parameters do not vary greatly, we concluded the MDP could be replaced with a simple implementation of the optimal strategy. To verify this hypothesis, we replaced the general MDP code with three simple lines of code implementing the $eA$ strategy, which we determined to be optimal for the particular parameters of the problem. Using log files recorded during the actual auctions reported in (Scerri, Pynadath, & Tambe, 2001), we experimentally verified that both the MDP and the $eA$ strategy produced the same result. Table 6 shows the percentage of available auction time remaining (e.g., if the auction was opened four days before the role should be performed, closing the auction one day before would correspond to 25%) when the MDP version and the $eA$ version of the code closed the auction. The number of bids is used to estimate the agent's expected decision quality. The timing of the auction closing is close, certainly within just a few hours. The result is not precisely the same for the MDP and strategy implementations, because the MDP implementation was more reactive to incoming bids than the strategy implementation.

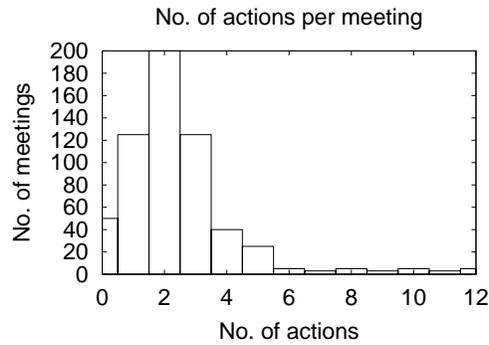

Figure 9: The frequency distribution of the number of steps taken in an AA strategy for the meeting scenario. If no actions were taken for a meeting, the meeting was cancelled before Friday started AA reasoning.

| Location | $A$ | $e$ | $eA$ | $e\mathcal{D}A$ | MDP |
|---|---|---|---|---|---|
| Small meeting, active participant | | | | | |
| office | 14.8 | -277 | 41.9 | 42.05 | $\mathcal{D}\mathcal{D}e\mathcal{D}A$ |
| not @ dept. | 14.8 | -6E7 | 31.4 | 28.0 | $\mathcal{D}\mathcal{D}eA$ |
| @ meet loc. | 14.8 | -2E5 | 39.2 | 39.1 | $eA$ |
| Large meeting, passive participant | | | | | |
| office | 14.6 | -7E12 | 30.74 | 30.65 | $\mathcal{D}\mathcal{D}eA$ |
| not @ dept. | 14.6 | -2E17 | 14.6 | 7.7 | $\mathcal{D}\mathcal{D}eA$ |
| @ meet loc. | 14.5 | -7E14 | 25.1 | 23.5 | $eA$ |

Table 5: EU values for the simple strategies as calculated from the model. The last column shows the strategy actually followed by the MDP.

| Date | No. Bids | MDP | $eA$ |
|------|----------|-----|------|
| 7/20/00 | 9 | 25% | 26% |
| 7/27/00 | 7 | 14% | 20% |
| 8/3/00 | 8 | 29% | 23% |

Table 6: Auction results. The "MDP" column shows the percentage of available auction time remaining when the MDP chose to close the auction. The "$eA$" column shows the percentage of available auction time remaining when the strategy $eA$, with $EQ_e^d(t)$ proportional to the number of bids received ("No. Bids" column), would have closed the auction.

To confirm that the need for strategies was not a phenomenon unique to the particular settings of the E-Elves, an experiment was run with randomly generated configurations of entities. The wait cost for each configuration increased exponentially, with the rate of accrual varying from configuration to configuration. The configurations contained between 3 and 25 entities, with randomly chosen Markovian response probabilities and randomly chosen, but constant, decision-making quality. The cost and value of a $\mathcal{D}$ action was also randomly selected. In each configuration, there was an agent that could respond instantly, but with lower decision quality than any of the other entities. For each configuration, the optimal transfer-of-control strategy was found. Figure 10(a) shows the percentage of optimal strategies (z-axis) that were of each length (y-axis "|Opt. Strat.|"), separated according to the rate at which wait costs accrued (x-axis, "Wait Cost Param"). The figure shows that if the rate at which the wait cost accrues is very low, most optimal strategies are of length one, with the agent just handing control to the entity with the highest decision-making quality. When the rate of wait cost accrual is high, most strategies are of length two, with the agent briefly giving the best decision maker an opportunity to make the decision but taking back control and acting before the wait costs became too high. For intermediate values of the wait cost parameter, there was considerably more variation in the length of the optimal strategy. Figure 10(b) shows the percentage of optimal strategies for each length when the wait cost parameter is 0.12 (i.e., a slice through Figure 10(a)). Hence, strategies often contained several transfers of control and several coordination changes. Thus, this experiment shows that complex transfer-of-control strategies are useful, not only in E-Elves, but in a range of domains, especially those in which wait costs are neither negligible nor accruing too fast.

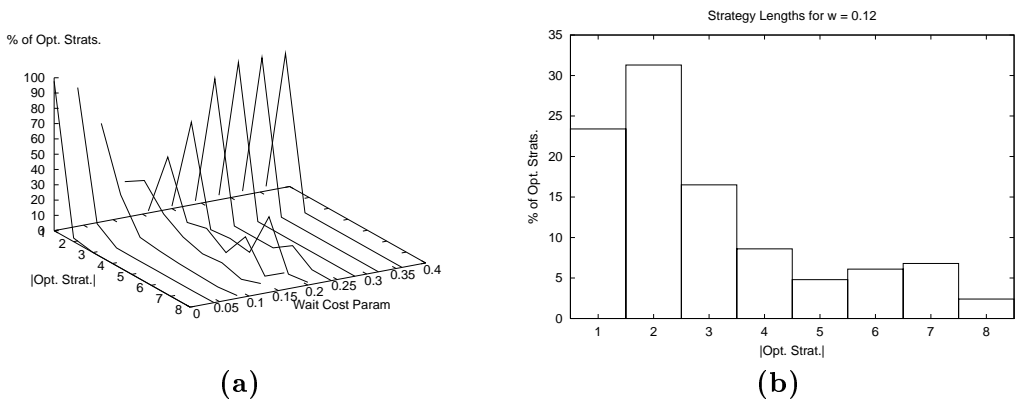

**(a)**                                                              **(b)**

Figure 10: (a) Percentage of optimal strategies having a certain length, broken down according to how fast wait costs are accruing. (b) Percentage of optimal strategies having certain length for wait cost parameter = 0.12.

Thus, we have shown that the MDP produces strategies and that Friday follows these strategies in practice. Moreover, the strategies followed are the ones predicted by the model. Of practical use, when we followed a prediction of the model, i.e., that an MDP was not required for auctions, we were able to substantially reduce the complexity of one part of the system. Finally, we showed that the need for strategies was not specifically a phenomenon of the E-Elves domain.

### 5.4 MDP Experiments

Experience using the MDP approach to AA in the E-Elves indicates that it is effective at making reasonable AA decisions. However, in order to determine whether MDPs are a generally useful tool for AA reasoning, more systematic experiments are required. In this section, we present such systematic experiments to determine important properties of MDPs for AA. The MDP reward function is designed to result in the optimal strategy being followed in each state.

In each of the experiments, we vary one of the $\lambda$ parameters that are the weights of the different factors in Equation 10. The MDP is instantiated with each of a range of values for the parameter and a policy produced for each value. In each case, the total policy is defined over 2800 states. The policy is analyzed to determine some basic properties of that policy. In particular, we counted the number of states in which the policy specifies to ask, to delay, to say the user is attending and to say the user is not attending. The statistics show broadly how the policy changes as the parameters change, e.g., whether Friday gives up autonomy more or less when the cost of a coordination change is increased. The first aim of the experiments is to simply confirm that policies change in the desired and expected way when parameters in the reward function are changed. For instance, if Friday's expected decision quality is increased, there should be more states where it makes an autonomous decision. Secondly, from a practical perspective it is critical to understand how sensitive the MDP policies are to small variations in parameters, because such sensitivity would mean that any small variations in parameter values can significantly impact MDP performance. Finally, the experiments reveal some interesting phenomena.

The first experiment looks at the effect of the $\lambda_1$ parameter from Equation 10, represented in the delay MDP implementation by the *team repair cost* (function $g$ from Equation 12), on the policies produced by the delay MDP. This parameter determines how averse Friday should be to changing coordination constraints. Figure 11 shows how some properties of the policy change as the team repair cost value is varied. The x-axis gives the value of the *team repair cost*, and the y-axis gives the number of times that action appears in the policy. Figure 11(a) shows the number of times Friday will ask the user for input. The number of times it will transfer control exhibits an interesting phenomenon: the number of asks has a maximum at an intermediate value for the parameter. For the low values, Friday can "confidently" (i.e., its decision quality is high) make decisions autonomously, since the cost of errors is low, hence there is less value to relinquishing autonomy. For very high team repair costs, Friday can "confidently" decide autonomously not to make a coordination change. It is in the intermediate region that Friday is uncertain and needs to call on the user's decision making more often. Furthermore, as the cost of delaying the meeting increases, Friday will delay the meeting less (Figure 11(b)) and tell the team the user is not attending more often (Figure 11(d)). By doing so, Friday gives the user less time to arrive at the meeting, choosing instead to just announce that the user is not attending. Essentially, Friday's decision quality has become close enough to the user's decision quality that asking the user is not worth the risk that they will not respond and the cost of asking for their input. Except for a jump between a value of zero and any non-zero value, the number of times Friday says the user is attending does not change (Figure 11(c)). The delay MDP in use in the E-Elves has the team repair cost parameter set at two. Around this value the policy changes little, hence slight changes in the parameter do not lead to large changes in the policy.

In the second experiment, we vary the $\lambda_2$ parameter from Equation 10, implemented in the delay MDP by the variable *team wait cost* (function $h$ from Equation 13). This is the factor that determines how heavily Friday should weigh differences between how the team expects the user will fulfill the role and how the user will actually fulfill the role. In particular, it determines the cost of having other team members wait in the meeting room for the user. Figure 12 shows the changes to the policy when this parameter is varied (again the x-axis shows the value of the parameter and the y-axis shows the number of times the action appears in the policy). The graph of the number of times the agent asks in the policy (Figure 12(a)), exhibits the same phenomena as when the $\lambda_1$ parameter was varied, i.e., increasing and then decreasing as the parameter increases. The graphs show that, as the cost of teammates' time increases, Friday acts autonomously more often (Figure 12(b-d)). Friday asks whenever the potential costs of asking are lower than the potential costs of errors it makes – as the cost of time waiting for a user decision increases, the balance tips towards acting. Notice that the phenomenon of the number of asks increasing then decreasing occurs in the same way that it did for the $\lambda_1$ parameter; however, it occurs for a slightly different reason. In this case, when waiting costs are low, Friday's decision-making quality is high so it acts autonomously. When the waiting costs are high, Friday cannot afford the risk that the user will not respond quickly, so it again acts autonomously (despite its decision quality being low). Figure 12(b) shows that the number of delay actions taken by Friday increases, but only in states in which the meeting has already been delayed twice. This indicates that the normally very expensive third delay of the same meeting starts to become worthwhile if the cost of having teammates wait in the meeting room is very high. In the delay MDP, a value of 1 is used for $\lambda_2$. The decision to transfer control (i.e., ask) is not particularly sensitive to changes in the parameter around this value—again, slight changes will not have a significant impact.

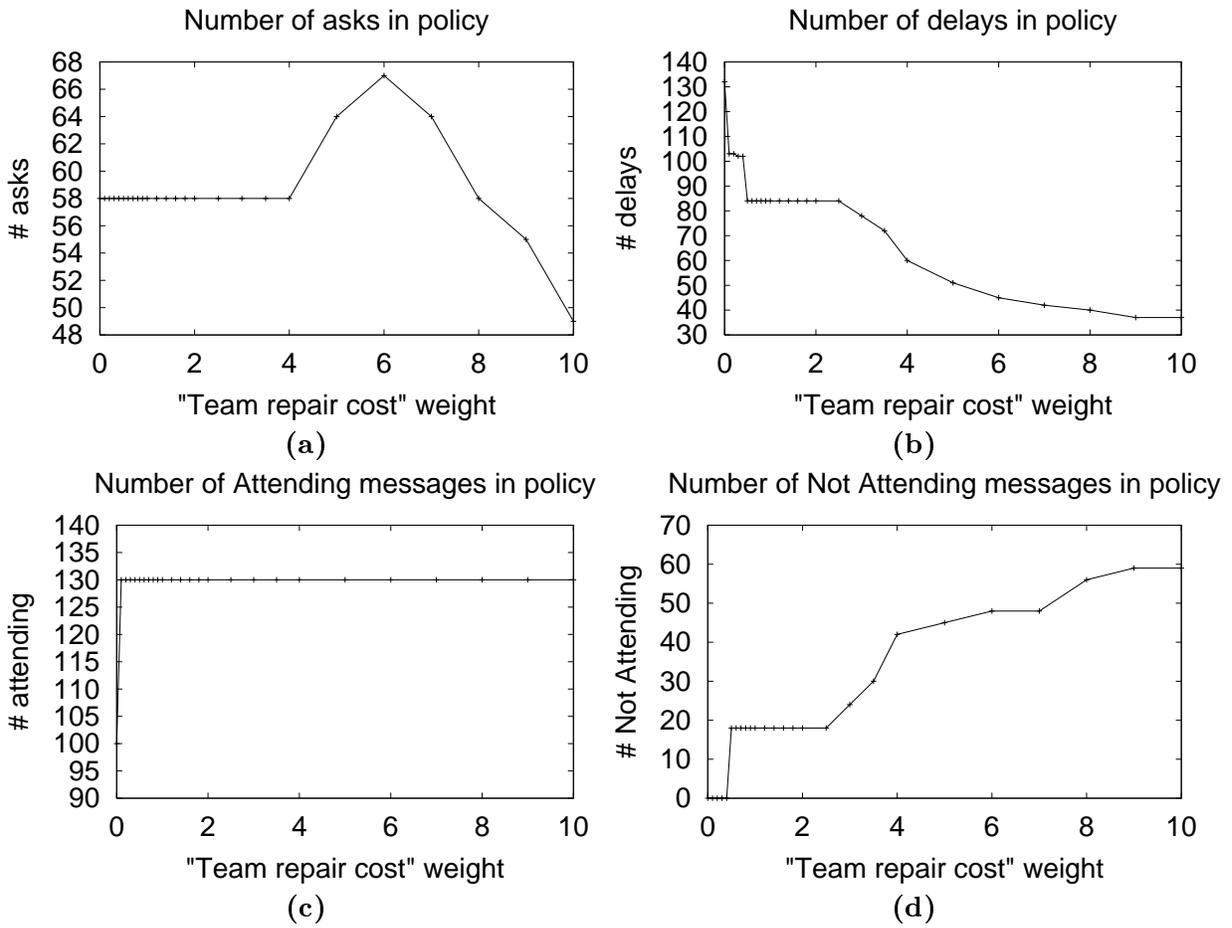

Figure 11: Properties of the MDP policy as team repair cost is varied.

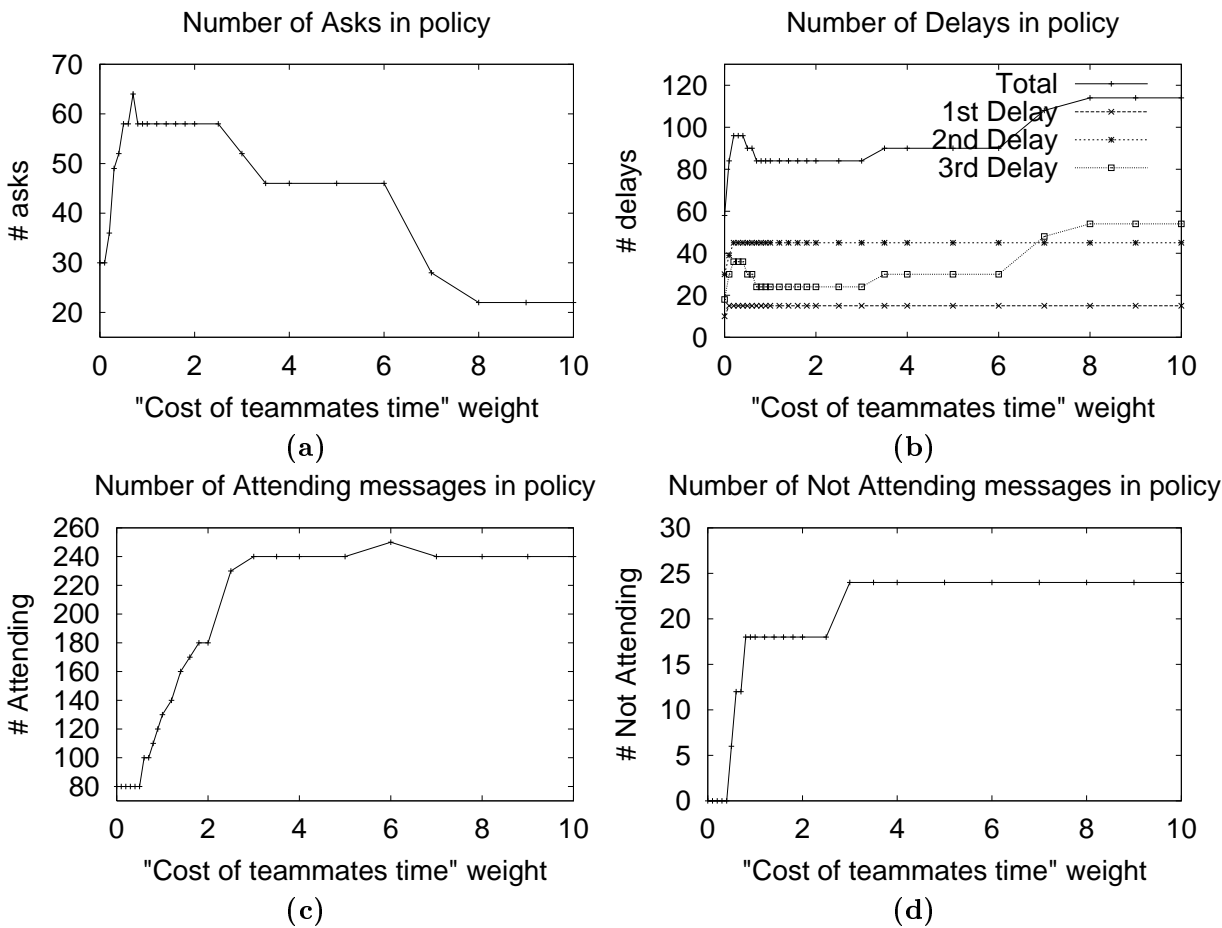

Figure 12: Properties of the MDP policy as teammate time cost is varied. (b) shows the number of times the meeting is delayed in states where it has not yet been delayed, where it has been delayed once already, and where it has been delayed twice already.

In the third experiment, the value of the $\lambda_3$, the weight of the joint task, was varied (Figure 13). In the E-Elves, the value of the joint task includes the value of the user to the meeting and the value of the meeting without the user. In this experiment, the value of the meeting without the user is varied. Figure 13 shows how the policy changes as the value of the meeting without the user changes (again the x-axis shows the value of the parameter and the y-axis shows the number of times the action appears in the policy). These graphs show significantly more instability than for the other $\lambda$ values. These large changes are a result of the simultaneous change in both the utility of taking key actions and the expected quality of Friday's decision making, e.g., the utility of saying the user is attending is much higher if the meeting has very low value without that user. In the current delay MDP, this value is set at 0.25, which is in a part of the graph that is very insensitive to small changes of the parameter.

In the three experiments above, the specific E-Elves parameters were in regions of the graph where small changes in the parameter do not lead to significant changes in the policy. However, there were regions of the graphs where the policy *did* change dramatically for small changes in a parameter. This indicates that in some domains, with parameters different to those in E-Elves, the policies will be sensitive to small changes in the parameters.

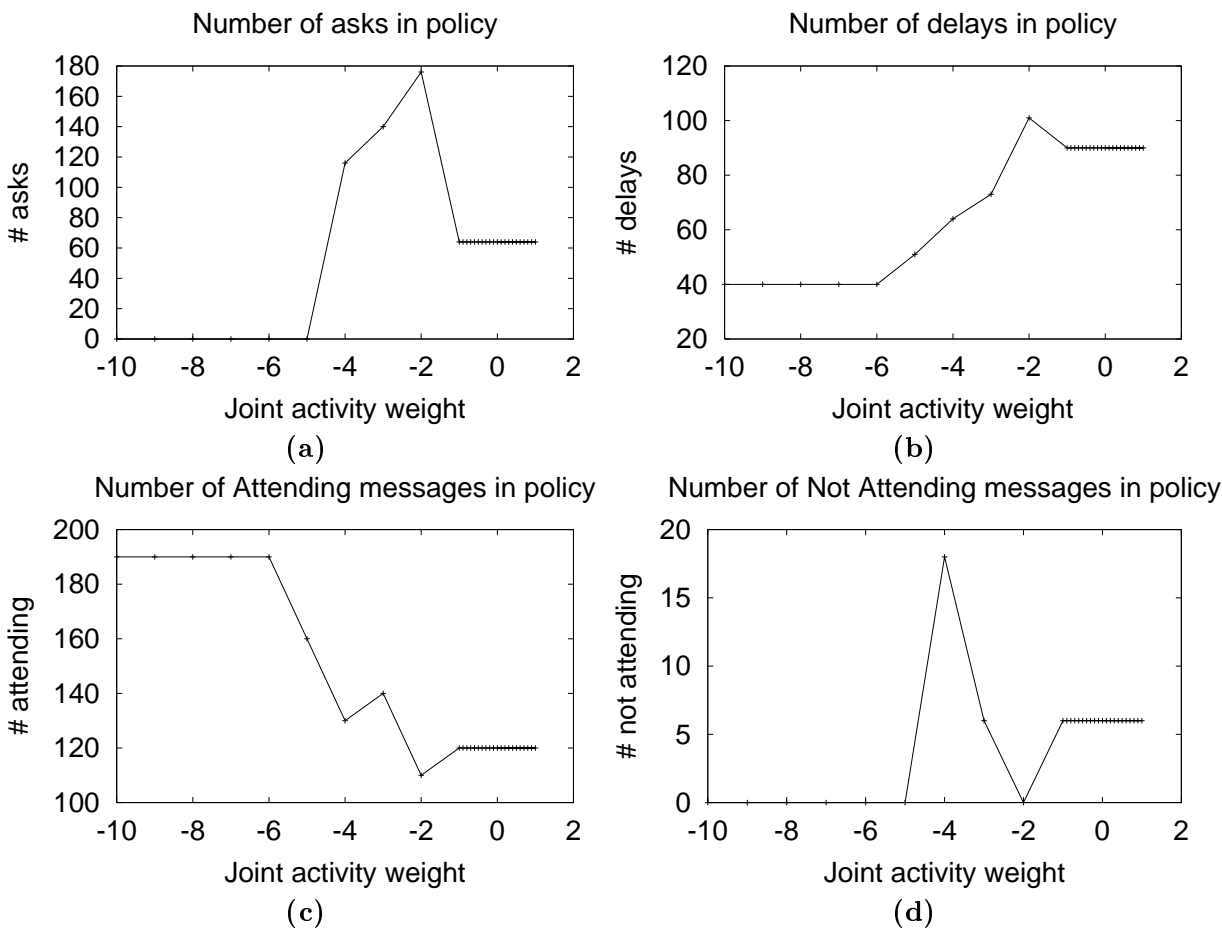

Figure 13: Properties of the MDP policy as the importance of a successful joint task is varied.

The above experiments show three important properties of the MDP approach to AA. First, changing the parameters of the reward function generally lead to the changes in the policy that are expected and desired. Second, while the value of the parameters influenced the policy, the effect on the AA reasoning was often reasonably small, suggesting that small errors in the model should not affect users too greatly. Finally, the interesting phenomena of the number of asks reaching a peak at intermediate values of the parameters was revealed.

The three previous experiments have examined how the behavior of the MDP changes as the parameters of the reward function are changed. In another experiment, a central domain-level parameter affecting the behavior of the MDP, i.e., the probability of getting a user response and the cost of getting that response (corresponding to $f_4$), is varied. Figure 14 shows how the number of times Friday chooses to ask (y-axis) varies with both the expected time to get a user response (x-axis) and the cost of doing so (each line on the graph represents a different cost). The MDP performs as expected, choosing to ask more often if the cost of doing so is low and/or it is likely to get a prompt response. Notice that, if the cost is low enough, Friday will sometimes choose to ask the user even if there is a long expected response time. Conversely, if the expected response time is sufficiently high, Friday will assume complete autonomy. This graph also shows that there is a distinct change in the number of asks at some point (depending on the cost), but outside this change point the graphs are relatively flat. The key reason for the fairly rapid change in the number of asks is that often the difference between the quality of Friday's and the user's decision making is in a fairly small range. As the mean response time increases, the expected wait costs increase, eventually becoming high enough for Friday to decide to act autonomously instead of asking.

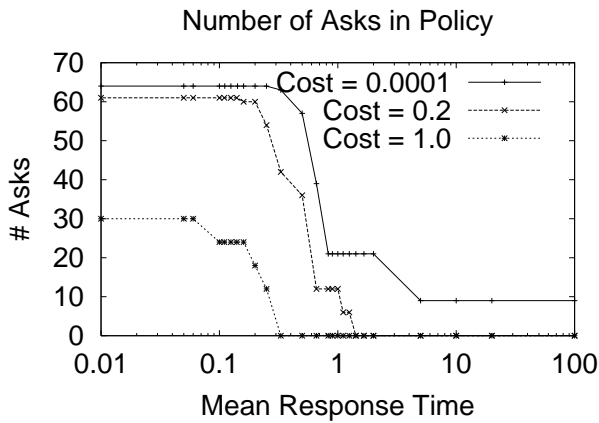

Figure 14: Number of ask actions in policy as the mean response time (in minutes) is varied. The x-axis uses a logarithmic scale.

We conclude this section with a quantitative illustration of the impact constraints have on strategy selection. In this experiment, we merged user-specified constraints from all the E-Elves users, resulting in a set of 10 distinct constraints. We started with an unconstrained instance of the *delay MDP* and added these constraints one at a time, counting the strategies that satisfied the applied constraints. We then repeated these experiments on expanded instances of the *delay MDP*, where we increased the initial state space by increasing the frequency of decisions (i.e., adding values to the *time-relative-to-meeting* feature). This expansion results in three new delay MDPs, which are artificial, but are influenced by the real delay MDP. Figure 15a displays these results (on a logarithmic scale), where line *A* corresponds to the original *delay MDP* (2760 states), and lines *B* (3320 states), *C* (3880 states), and *D* (4400 states) correspond to the expanded instances. Each data point is a mean over five different orderings of constraint addition. For all four MDPs, the constraints substantially reduce the space of possible agent behaviors. For instance, in the original *delay MDP*, applying all 10 constraints eliminated 1180 of the 2760 original states from consideration, and reduced the mean number of viable actions per acceptable state from 3.289 to 2.476. The end result is a 50% reduction in the size ($\log_{10}$) of the strategy space. On the other hand, constraints alone did not provide a complete strategy, since all of the plots stay well above 0, even with all 10 constraints. Since none of the individual users were able/willing to provide 10 constraints, we cannot expect anyone to add enough constraints to completely specify an entire strategy. Thus, the MDP representation and associated policy selection algorithms are still far from redundant.

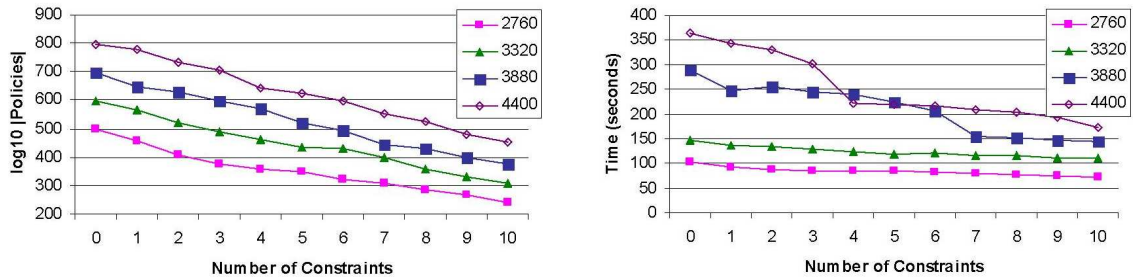

Figure 15: (a) Number of possible strategies (logarithmic). (b) Time required for strategy generation.

The constraints' elimination of behaviors also decreases the time required for strategy selection. Figure 15b plots the total time for constraint propagation and value iteration over the same four MDPs as in Figure 15a (averaged over the same five constraint orderings). Each data point is also a mean over five separate iterations, for a total of 25 iterations per data point. The values for the zero-constraint case correspond to standard value iteration without constraints. The savings in value iteration over the restricted strategy space dramatically outweigh the cost of pre-propagating the additional constraints. In addition, the savings increase with the size of the MDP. For the original *delay MDP* (*A*), there is a 28% reduction in policy-generation time, while for the largest MDP (*D*), there is a 53% reduction. Thus, the introduction of constraints can provide dramatic acceleration of the agent's strategy selection.

## 6. Related Work

We have discussed some related work in Section 1. This section adds to that discussion. In Section 6.1, we examine two representative AA systems – where detailed experimental results have been presented – and explain those results via our model. This illustrates the potential applicability of our model to other systems. In Section 6.2, we examine other AA systems and other areas of related work, such as meta-reasoning, conditional planning and anytime algorithms.

### 6.1 Analyzing Other AA Work Using the Strategy Model

Goodrich, Olsen, Crandall, and Palmer (2001) report on tele-operated teams of robots, where both the user's high-level reasoning and the robots' low-level skills are required to achieve some task. Within this domain, they have examined the effect of *user neglect* on robot performance. The idea of user neglect is similar to our idea of entities taking time to make decisions; in this case, if the user "neglects" the robot, the joint task takes longer to perform. In this domain, the coordination constraint is that user input must arrive so that the robot can work out the low-level actions it needs to perform. Four control systems were tested on the robot, each giving a different amount of autonomy to the robot, and the performance was measured as user neglect was varied.

Although quite distinct from the E-Elves system, mapping Goodrich's team of robots to our AA problem formulation provides some interesting insights. This system has the interesting feature that the entity the robot can call on for a decision, i.e., the user, is also part of the team. Changing the autonomy of the robot effectively changes the nature of the coordination constraints between the user and robot. Figure 16 shows the performance (y-axis) of the four control policies as the amount of user neglect was increased (x-axis). The experiments showed that higher robot autonomy allowed the operator to "neglect" the robot more without as serious an impact on its performance.

The notion of transfer-of-control strategies can be used to qualitatively predict the same behavior as was observed in practice, even though Goodrich et al. (2001) did not use the notion of strategies. The lowest autonomy control policy used by Goodrich et al. (2001) was a pure tele-operation one. Since the robot cannot resort to its own decision making, we represent this control policy with a strategy $U$, i.e., control indefinitely in the hands of the user. The second control policy allows the user to specify *waypoints* and on-board intelligence works out the details of getting to the waypoints. Since the robot has no high-level decision-making ability, the strategy is simply to give control to the user. However, since the coordination between the robot and user is more abstract, i.e., the coordination constraints are looser, the wait cost function is less severe. Also the human is giving less detailed guidance than in the fully tele-operated case (which is not as good according to (Goodrich et al., 2001)), hence we use a lower value for the expected quality of the user decision. We denote this approach $U_wp$ to distinguish it from the fully tele-operated case. The next control policy allows the robot to choose its own waypoints given that the user inputs *regions of interest*. The robot can also accept waypoints from the user. The ability for the robot to calculate waypoints is modeled as a $\mathcal{D}$, since it effectively changes the coordination between the entities, by removing the user's need to give waypoints. We model this control policy as the strategy $U\mathcal{D}U$. The final control policy is full autonomy, i.e., $A$.

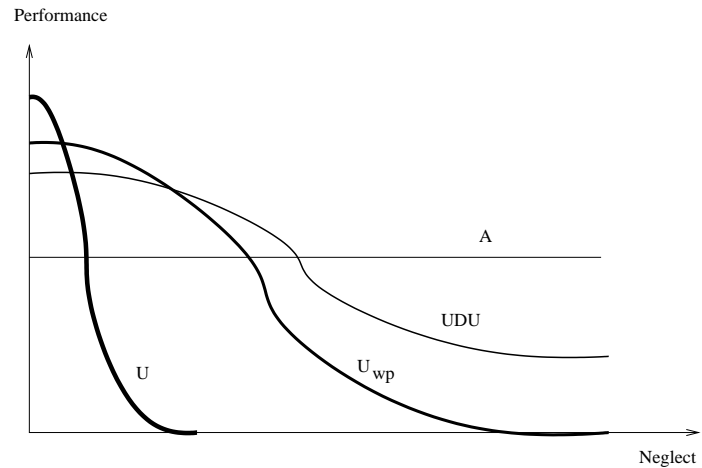

**(a)**

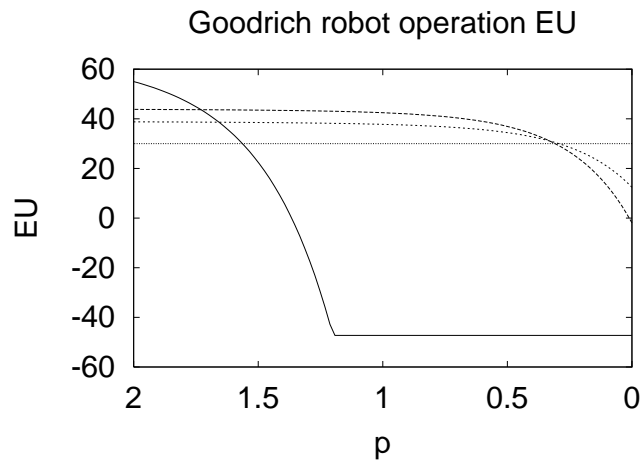

**(b)**

Figure 16: Goodrich at al's various control strategies plotted against neglect. (a) Experimental results. Thinner lines represent control systems with more intelligence and autonomy. (b) Results theoretically derived from model of strategies presented in this article (p is the parameter to the probability of response function).

Robot decision making is inferior to that of the user, hence the robot's decision quality is less than the user's. The graphs of the four strategies, plotted against the probability of response parameter (getting smaller to the right, to match "neglect" in the Goodrich et al graph) is shown in Figure 16. Notice that the shape of the graph theoretically derived from our model, shown in Figure 16(b), is qualitatively the same as the shape of the experimentally derived graph, Figure 16(a). Hence, the theory predicted qualitatively the same performance as was found from experimentation.

A common assumption in earlier AA work has been that if any entity is asked for a decision it will make that decision promptly, hence strategies handling the contingency of a lack of response have not been required. For example, Horvitz's (1999) work using decision theory is aimed at developing general, theoretical models for AA reasoning for a user at a workstation. A prototype system, called LookOut, for helping users manage their calendars has been implemented to test these ideas (Horvitz, 1999). Although such systems are distinctly different from E-Elves, mapping them to our problem formulation allows us to analyze the utility of the approaches across a range of domains without having to implement the approach in those domains.

A critical difference between Horvitz's work and our work is that LookOut does not address the possibility of not receiving a (timely) response. Thus, complex strategies are not required. In the typical case for LookOut, the agent has three options: to take some action, not to take the action, or to engage in dialog. The central factor influencing the decision is whether the user has a particular goal that the action would aid, i.e., if the user has the goal, then the action is useful, but if he/she does not have the goal, the action is disruptive. Choosing to act or not to act corresponds to pursuing strategy $A$.[7] Choosing to seek user input corresponds to strategy $U$. Figure 17(a) shows a graph of the different options plotted against the probability the user has the goal (corresponds to Figure 6 in Horvitz (1999)). The agent's expected decision quality, $EQ_A^d(t)$ is derived from Equation 2 in Horvitz (1999). (In other words, Horvitz's model performs more detailed calculations of expected decision quality.) Our model then predicts the same selection of strategies as Horvitz does, i.e., choosing strategy $A$ when $EQ_A^d(t)$ is low, $U$ otherwise (assuming that only those two strategies are available). However, our model further predicts something that Horvitz did not consider, i.e., that if the rate at which wait costs accrue becomes non-negligible then the choice is not as simple. Figure 17(b) shows how the EU of the two strategies changes as the rate of wait costs accruing is increased. The fact that the optimal strategy varies with wait cost suggests that Horvitz's approach would not immediately be appropriate for a domain where wait costs were non-negligible, e.g., it would need to be modified in many multi-agent settings.

## 6.2 Other Approaches to AA

Several different approaches have been taken to the core problem of whether and when to transfer decision-making control. For example, Hexmoor examines how much time the agent has to do AA reasoning (Hexmoor, 2000). Similarly, in the Dynamic Adaptive Autonomy framework, a group of agents allocates votes amongst themselves, hence defining the amount of influence each agent has over a decision and thus, by their definition, the autonomy of the agent with respect to the decision (Barber, Martin, & Mckay, 2000b). For the related application of meeting scheduling Cesta, Collia, and D'Aloisi (1998) have taken the approach of providing powerful tools for users to constrain and monitor the behavior of their proxy agents, but the agents do not explicitly reason about relinquishing control to the user. While at least some of this work is done in a multiagent context, the possibility of multiple transfers of control is not considered.

Complementing our work, other researchers have focused on issues of architectures for AA. For instance, an AA interface to the 3T architecture (Bonasso, Firby, Gat, Kortenkamp,

---

7. We consider choosing not to act an autonomous decision, hence categorize it in the same way as autonomous action

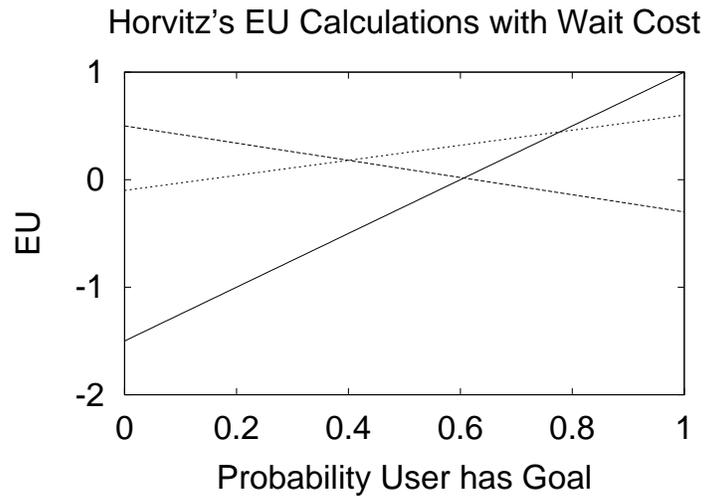

(a)

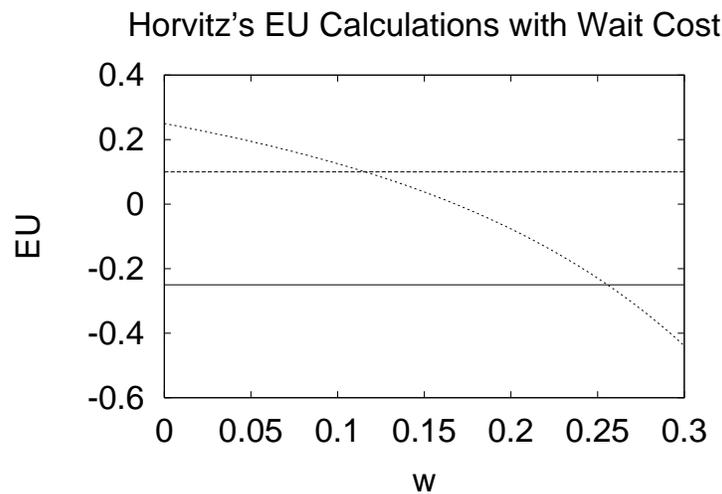

(b)

Figure 17: EU of different agent options. The solid (darkest) line shows the EU taking an autonomous action, the dashed (medium dark) line shows the EU of autonomously deciding not to act and the dotted line shows the EU of transferring control to the user. (a) Plotted against the probability of user having goal, no wait cost. (b) plotted against wait cost, fixed probability of user having goal.

Miller, & Slack, 1997) has been implemented to solve human-machine interaction problems experienced in a number of NASA projects (Brann, Thurman, & Mitchell, 1996). The experiences showed that interaction with the system was required all the way from the deliberative layer through to detailed control of actuators. The AA controls at all layers are encapsulated in what is referred to as the 3T's fourth layer – the interaction layer (Schreckenghost, 1999). A similar area where AA technology is required is for safety-critical intelligent software, such as for controlling nuclear power plants and oil refineries (Musliner & Krebsbach, 1999). That work has resulted in a system called AEGIS (Abnormal Event Guidance and Information System) that combines human and agent capabilities for rapid reaction to emergencies in a petro-chemical refining plant. AEGIS features a *shared task representation* that both the users and the intelligent system can work with (Goldman, Guerlain, Miller, & Musliner, 1997). A key hypothesis of the work is that the model needs to have multiple levels of abstraction so that the user can interact at the level they see fit. Interesting work by Fong, Thorpe, and Baur (2002) has extended the idea of tele-operated robotics by re-defining the relationship between the robot and user as a collaborative one, rather than the traditional master-slave configuration. In particular, the robot treats the human as a resource that can perform perceptual or cognitive functions that the robot determines it cannot adequately perform. However, as yet the work has not looked at the possibility that the user is not available to provide input when required, which would require the robot perform more complex transfer-of-control reasoning.

While most previous work in AA has ignored complex strategies for AA, there is work in other research fields that is potentially relevant. For example, the research issues addressed by fields such as mixed-initiative decision-making (Collins, Bilot, Gini, & Mobasher, 2000b), anytime algorithms (Zilberstein, 1996), multi-processor scheduling (Stankovic, Ramamritham, & Cheng, 1985), meta-reasoning (Russell & Wefald, 1989), game theory (Fudenberg & Tirole, 1991), and contingency plans (Draper, Hanks, & Weld, 1994; Peot & Smith, 1992) all have, at least superficial, similarities with the AA problem. However, it turns out that the core assumptions and focus of these other research areas are different enough that the algorithms developed in these related fields are not directly applicable to the AA problem.

In mixed-initiative decision making a human user is assumed to be continually available (Collins et al., 2000b; Ferguson & Allen, 1998), negating any need for reasoning about the likelihood of response. Furthermore, there is often little or no time pressure or coordination constraints. Thus, while the basic problem of transferring control between a human and agent is common to both mixed-initiative decision making and AA, the assumptions are quite different leading to distinct solutions. Likewise, other related research fields make distinctly different assumptions which lead to distinctly different solutions. For instance, contingency planning (Draper et al., 1994; Peot & Smith, 1992) deals with the problem of creating plans to deal with critical developments in the environment. Strategies are related to contingency planning in that they are plans to deal with the specific contingency of an entity not making a decision in a manner that maintains coordination. However, in contingency planning, the key difficulty is in creating the plans. In contrast, in AA, creating strategies is straightforward and the key difficulty is choosing between those strategies. Our contribution is in recognizing the need for strategies in addressing the AA problem, instantiating such strategies via MDPs, and the development of a general, domain-independent reward function that leads to an MDP choosing the optimal strategy for a particular situation.

Similarly, another related research area is meta-reasoning (Russell & Wefald, 1989). Meta-reasoning work looks at online reasoning about computation. A type of meta-reasoning, most closely related to AA, chooses between sequences of computations with different ex-pected quality and running time, subject to the constraint that choosing the highest-quality sequence of computations is not possible (because it takes too long) (Russell & Wefald, 1989). The idea is to treat computations as actions and "meta-reason" about the EU of doing certain combinations of computation and (base-level) actions. The output of meta-reasoning is a sequence of computations that are executed in sequence. AA parallels meta-reasoning if we consider reasoning about transferring control to entities as reasoning about selecting computations, i.e., we think of entities as computations. However, in AA, the aim is to have one entity make a high-quality decision, while in meta-reasoning, the aim is for a sequence of computations to have some high quality. Moreover, the meta-reasoning assumption that computations are guaranteed to return an immediate decision, does not apply in AA. Finally, meta-reasoning looks for a sequence of computations that use a fixed amount of time, while AA reasons about trading off extra time for a better decision (possibly buying time with a $\mathcal{D}$ action). Thus, algorithms developed for meta-reasoning are not applicable to AA.

Another research area with conceptual similarity to AA is the field of *anytime algorithms* (Zilberstein, 1996). An anytime algorithm quickly finds an initial solution and then incrementally tries to improve the solution until stopped. The AA problem is similar when we assume that the agent itself can make an immediate decision, because the problem then has the property that a solution is always available (an important property of an anytime algorithm). However, this will not be the case in general, i.e., the agent will not always have an answer. Furthermore, anytime algorithms do not generally need to deal with multiple, distributed entities, nor do they have the opportunity to change coordination (i.e., using a $\mathcal{D}$ action).

Multi-processor scheduling looks at assigning tasks to nodes in order to meet certain time constraints (Stankovic et al., 1985). If entities are thought of as "nodes", then AA is also about assigning tasks to nodes. In multiprocessor scheduling, the quality of the computation performed on each of the nodes is usually assumed to be equal, i.e., the nodes are homogeneous. Thus, reasoning that trades off quality and time is not required, as it is in AA. Moreover, deadlines are externally imposed for multi-processor scheduling algorithms, rather than being flexibly reasoned about as in AA. Multi-processor scheduling algorithms can sometimes deal with a node rejecting a task because it cannot fulfill the time constraints or network failures. However, while the AA problem focuses on failure to get a response as a central issue and load balancing as an auxiliary issue, multi-processor scheduling has the opposite focus. The difference in focus leads to algorithms being developed in the multiprocessor scheduling community that are not well suited to AA (and vice versa).

## 7. Conclusions

Adjustable autonomy is critical to the success of real-world agent systems because it allows an agent to leverage the skills, resources and decision-making abilities of other entities, both human and agent. Previous work has addressed AA in the context of single-agent and single-human scenarios, but those solutions do not scale to increasingly complex multi-agent systems. In particular, previous work used rigid, one-shot transfers of control that did not consider team costs and, more importantly, did not consider the possibility of costly miscoordination between team members. Indeed, when we applied a rigid transfer-of-control approach to a multi-agent context, it failed dramatically.

This article makes three key contributions to enable the application of AA in more complex multiagent domains. First, the article introduces the notion of a transfer-of-control strategy. A transfer-of-control strategy consists of a conditional sequence of two types of actions: (i) actions to transfer decision-making control and (ii) actions to change an agent's pre-specified coordination constraints with team members, aimed at minimizing miscoordination costs. Such strategies allow agents to plan sequences of transfer-of-control actions. Thus, a strategy allows the agent to transfer control to entities best able to make decisions, buy more time for decisions to be made and still avoid miscoordination — even if the entity to which control is transferred fails to make the decision. Additionally, we introduced the idea of changing coordination constraints as a mechanism for giving the agent more opportunity to provide high-quality decisions, and we showed that such changes can, in some cases, be an effective way of increasing the team's expected utility.

The second contribution of this article is a mathematical model of AA strategies that allows us to calculate the expected utility of such strategies. The model shows that while complex strategies are indeed better than single-shot strategies in some situations, they are not always superior. In fact, our analysis showed that no particular strategy dominates over the whole space of AA decisions; instead, different strategies are optimal in different situations.

The third contribution of this article is the operationalization of the notion of transfer-of-control strategies via Markov Decision Processes and a general reward function that leads the MDP to find optimal strategies in a multiagent context. The general, domain-independent reward function should allow our approach to potentially be applied to other multi-agent domains. We implemented, applied, and tested our MDP approach to AA reasoning in a real-world application supporting researchers in their daily activities. Daily use showed the MDP approach to be effective at balancing the need to avoid risky autonomous decisions and the potential for costly miscoordination. Furthermore, detailed experiments showed that the policies produced by the MDPs have desirable properties, such as transferring control to the user less often when the probability of getting a timely response is low. Finally, practical experience with the system revealed that users require the ability to manipulate the AA reasoning of the agents. To this end, we introduced a constraint language that allows the user to limit the range of behavior the MDP can exhibit. We presented an algorithm for processing such constraints, and we showed it to have the desirable property of reducing the time it takes to find optimal policies.

## 8. Future Work

The model of AA presented in this article is sufficiently rich to model a wide variety of interesting applications. However, there are some key factors that are not modeled in the current formulation that are required for some domains. One key issue is to allow an agent to factor the AA reasoning of other agents into its own AA reasoning. For instance, in the Elves domain, if one agent is likely to decide to delay a meeting, another agent may wait until that decision and avoid asking its user. Conversely, if an agent about to take back control of a decision knows another agent is going to continue waiting for user input, it might also continue to wait for input. Such interactions will substantially increase the complexity of the reasoning an agent needs to perform. In this article, we have assumed that the agent is finding a transfer-of-control strategy for a single, isolated decision. In general, there will be many decisions to be made at once and the agent will not be able to ignore the interactions between those decisions. For example, transferring control of many decisions to a user, reduces the probability of getting a prompt response to any of them. Reasoning about these interactions will add further complexity to the required reasoning of the agent.

Another focus of future work will be generalizing the AA decision making to allow other types of constraints — not just coordination constraints — to be taken into account. This would in turn require generalization of the concept of a $\mathcal{D}$ action to include other types of stop-gap actions and may lead to different types of strategies an agent could pursue. Additionally, transfer-of-control actions could be generalized to allow parts of a decision to be transferred, e.g., to allow input to be received from a user without transferring total control to him/her, or allow actions that could be performed collaboratively. Similarly, if actions were reversible, the agent could make the decision but allow the user to reverse it. We hope that such generalizations would improve the applicability of our adjustable autonomy research in more complex domains.

## Acknowledgments

This research was supported by DARPA award no. F30602-98-2-0108. The effort is being managed by Air Force Research Labs/Rome site. This article unifies, generalizes, and significantly extends approaches described in our previous conference papers (Scerri et al., 2001; Scerri, Pynadath, & Tambe, 2002; Pynadath & Tambe, 2001). We thank our colleagues, especially, Craig Knoblock, Yolanda Gil, Hans Chalupsky and Tom Russ for collaborating on the Electric Elves project. We would also like to thank the JAIR reviewers for their useful comments.

## Appendix A: An Example Instantiation of the Model

In this Appendix, we present a detailed look at one possible instantiation of the AA model. We use that instantiation to calculate the EU of commonly used strategies and show how that EU varies with parameters such as the rate of wait cost accrual and the time at which transfers of control are performed. In this instantiation, the agent, $A$, has only one entity to call on for a decision (i.e., the user $U$), hence $E = \{A, U\}$. For $\mathcal{W}(t)$, we use the following function:

$$\mathcal{W}(t) = \begin{cases} \omega \exp^{\omega t} & t \leq \vartriangleleft \\ \omega \exp^{\omega \vartriangleleft} & \text{otherwise} \end{cases} \tag{15}$$

The exponential wait cost function reflects the idea that a big delay is much worse than a small one. A polynomial or similar function could have also been used but an exponential was used since it makes the mathematics cleaner. For the probability of response we use: $P_{\top}(t) = \rho \exp^{-\rho t}$. A Markovian response probability reflects an entity that is just as likely to respond at the next point in time as they were at the previous point. For users moving around a dynamic environment, this turns out to be a reasonable approximation. The entities' decision-making quality is constant over time, in particular, $EQ_A^d(t) = \alpha$ and for $EQ_U^d(t) = \beta$. Assuming constant decision-making quality will not always be accurate in a dynamic environment since information available to an entity may change (hence influencing their ability to make the decision) however, for decisions involving static facts or preferences decision-making quality will be relatively constant. The functions are a coarse approximation of a range of interesting applications, including the E-Elves. Table 7 shows the resulting instantiated equations for the simple strategies (For convenience we let $\delta = \rho - \omega$). Figures 18(a) and (b) show graphically how the EU of the $eA$ strategy varies along different axes (w is the parameter to the wait cost function, higher w means faster accruing wait costs and p is the parameter to the response probability function, higher p means faster response). Notice how the EU depends on the transfer time (T) as much as it does on $\beta$ (the user's decision quality). Figure 18(d) shows the value of a $\mathcal{D}$ (as discussed earlier).

Figure 18(c) compares the EU of the $e\mathcal{D}eA$ and $e$ strategies. The more complex the transfer-of-control strategy (i.e., the more transfers of control it makes), the flatter the EU graph when plotted against wait cost (w) and response probability (p) parameters. In particular, the fall-off when the wait costs are high and the probability of response low is not so dramatic for the more complex strategy.

## Appendix B: Constraint Propagation Algorithm and its Correctness

In Section 4.4, we examined the need for user-specified constraints in conjunction with our MDP-based approach to strategies. We must thus extend the standard MDP policy evaluation algorithms to support the evaluation of strategies while accounting for both the standard quantitative reward function *and* these new qualitative constraints. This appendix provides the novel algorithm that we developed to evaluate strategies while accounting for

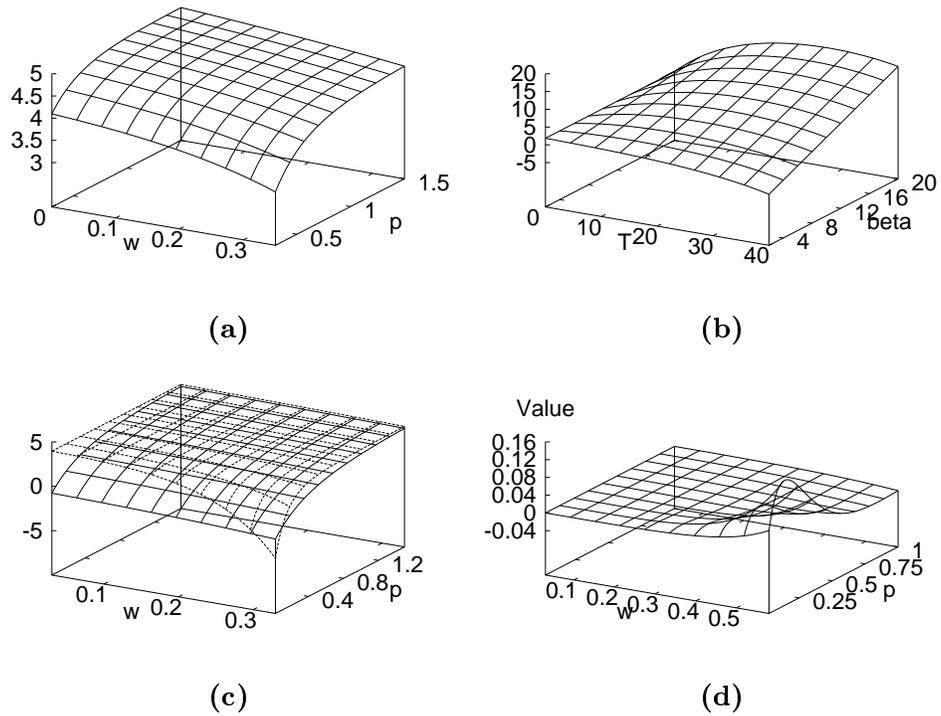

**(a)**                 **(b)**

**(c)**                 **(d)**

Figure 18: Equation 17, i.e., strategy $eA$ plotted against (a) $\omega$ (i.e., w, the rate at which wait costs accrue) and $\rho$ (i.e., p the likelihood of response) and (b) $T$ (transfer time) and $beta$ (the user's decision quality). (c) Comparing strategies $e\mathcal{D}eA$ and $e$ (dotted line is $e$). (d) The value of a $\mathcal{D}$.

$$EU_e^d t = \exp^{-\lhd\delta} \times \omega(\frac{\rho}{\delta} - 1) - \frac{\rho\omega}{\delta} + \beta \qquad (16)$$

$$EU_{eA}^d t = \omega \exp^{-T\delta}(\frac{\rho}{\delta} - 1) + \exp^{-\rho T}(\alpha - \beta) - \frac{\rho\omega}{\delta} + \beta \qquad (17)$$

$$EU_{e\mathcal{D}e A}^d t = \qquad\qquad\qquad\qquad\qquad\qquad\qquad\qquad\qquad\qquad (18)$$
$$\frac{\rho\omega}{\delta}(exp^{-\Delta\delta} - 1) + \beta(1 - \exp^{-\Delta\rho}) + \frac{\rho\omega \exp^{-\omega\mathcal{D}_{value}}}{\delta}(\exp^{-T\delta} - \exp^{-\Delta\delta}) +$$
$$(\mathcal{D}_{cost} - \beta)(\exp^{-\rho T} - \exp^{-\rho\Delta}) + \omega\exp^{\Delta\omega}(\exp^{-\omega\mathcal{D}_{value}} - 1)(\exp^{-\rho\Delta} - \exp^{-\rho T}) -$$
$$\exp^{-\rho T}(\mathcal{D}_{cost} - \alpha + \omega(\exp^{\omega\Delta} - \exp^{\omega(\Delta - \mathcal{D}_{value})} + \exp^{\omega(T - \mathcal{D}_{value})}))$$

Table 7: Instantiated AA EU equations for simple transfer of control strategies.

both. We also present a detailed proof that our algorithm's output is the correct strategy (i.e., the strategy with the highest expected utility, subject to the user-specified constraints).

In the standard MDP value iteration algorithm, the value of a strategy in a particular state is a single number, an expected utility $U$. With the addition of our two types of constraints, this value is now a tuple $\langle F, N, U \rangle$. $F$ represents a strategy's ability to satisfy the forbidding constraints; therefore, it is a boolean indicating whether the state is forbidden or not. $N$ represents a strategy's ability to satisfy the necessary constraints; therefore, it is the set of requiring constraints that will be satisfied. As in traditional value iteration, $U$ is the expected reward. For instance, if the value of a state, $V(s) = \langle \text{true}, \{c_{rs}\}, 0.3 \rangle$, then executing the policy from state $s$ will achieve an expected value of 0.3 and will satisfy required-state constraint $c_{rs}$. However, it is not guaranteed to satisfy any other required-state, nor any required-action, constraints. In addition, $s$ is forbidden, so there is a nonzero probability of violating a forbidden-action or forbidden-state constraint. We do not record *which* forbidding constraints the policy violates, since violating any one of them is equally bad. We *do* have to record which requiring constraints the policy satisfies, since satisfying all such constraints is preferable to satisfying only some of them. Therefore, the size of the value function grows linearly with the number of requiring constraints, but is independent of the number of forbidding constraints.

Following the form of standard value iteration, we initialize the value function over states by considering the immediate value of the strategy in the given state, without any lookahead. More precisely:

$$V^0(s) \leftarrow \left\langle \bigvee_{c \in C_{fs}} c(s), \{c \in C_{rs} | c(s)\}, R_S(s) \right\rangle \qquad (19)$$

Thus, the state $s$ is forbidden if any forbidden-state constraints immediately apply, and it satisfies those required-state constraints that immediately apply. As in standard value iteration, the expected utility is the value of the reward function in the state.

In value iteration, we must define an updated value function $V^{t+1}$ as a refinement of the previous iteration's value function, $V^t$. States become forbidden in $V^{t+1}$ if they violate any constraints directly or if *any* of their successors are forbidden according to $V^t$. States satisfy requirements if they satisfy them directly or if *all* of their successors satisfy the requirement. To simplify the following expressions, we define $S'$ to be the set of all successors: $\{s' \in S | M^a_{ss'} > 0\}$. The following expression provides the precise definition of this iterative step:

$$V^{t+1}(s) \leftarrow \max_{a \in A} \left\langle \bigvee_{c \in C_{fs}} c(s) \vee \bigvee_{c \in C_{fa}} c(s,a) \vee \bigvee_{V^t(s')=\langle F',N',U'\rangle, s' \in S'} F', \right.$$

$$\{c \in C_{rs}|c(s)\} \cup \{c \in C_{ra}|c(s,a)\} \cup \bigcap_{V^t(s')=\langle F',N',U'\rangle, s' \in S'} N',$$

$$\left. R_S(s) + R(s,a) + \sum_{V^t(s')=\langle F',N',U'\rangle, s' \in S'} M^a_{ss'} U' \right\rangle \tag{20}$$

Just as in standard value iteration, this iterative step specifies a maximization over all possible choices of action. However, with our two additional components to represent the value of the strategy with respect to the constraints, we no longer have an obvious comparison function to use when evaluating candidate actions. Therefore, we perform the maximization using the following preference ordering, where $x \prec y$ means that $y$ is preferable to $x$:

$$\langle t, N, U \rangle \prec \langle f, N', U' \rangle$$
$$\langle F, N, U \rangle \prec \langle F, N' \supset N, U' \rangle$$
$$\langle F, N, U \rangle \prec \langle F, N, U' > U \rangle$$

In other words, satisfying a forbidden constraint takes highest priority, satisfying more requiring constraints is second, and increasing expected value is last. We define the optimal action, $P(s)$, as the action, $a$, for which the final $V(s)$ expression above is maximized.

Despite the various set operations in Equation 20, the time complexity of this iteration step exceeds that of standard value iteration by only a linear factor, namely the number of constraints, $|C_{fs}| + |C_{fa}| + |C_{rs}| + |C_{ra}|$. The efficiency derives from the fact that the constraints are satisfied/violated independently of each other. The determination of whether a single constraint is satisfied/violated requires no more time than that of standard value iteration, hence the overall linear increase in time complexity.

Because expected value has the lowest priority, we can separate the iterative step of Equation 20 into two phases: constraint propagation and value iteration. During the constraint-propagation phase, we compute only the first two components of our value function, $\langle F, N, \cdot \rangle$. The value-iteration phase computes the third component, $\langle \cdot, \cdot, U \rangle$, as in standard value iteration. However, we can ignore any state/action pairs that, according to the results of constraint propagation, violate a forbidding constraint ($\langle t, N, \cdot \rangle$) or requiring constraint ($\langle f, N \subset C_{rs} \cup C_{ra}, \cdot \rangle$). Because of the component-wise independence of Equation 20, the two-phase algorithm computes an identical value function as the original, single-phase version (over state/action pairs that satisfy all constraints).

In the rest of this Appendix we provide a proof of the correctness of the modified value iteration policy. Given a policy, $P$, constructed according to the above algorithm, we must show that an agent following $P$ will obey the constraints specified by the user. If the agent begins in some state, $s \in S$, we must prove that it will satisfy all of its constraints if and only if $V(s) = \langle f, C_{ra} \cup C_{rs}, U \rangle$. We prove the results for forbidding and requiring constraints separately.

**Theorem 1** *An agent following policy, $P$, with value function, $V$, generated as in Section 4.4, from any state $s \in S$ will violate a forbidding constraint with probability zero if and only if $V(s) = \langle f, N, U \rangle$ (for some $U$ and $N$).*

**Proof:** We prove the theorem by induction over subspaces of the states, classified by how "close" they are to violating a forbidding constraint. More precisely, we partition the state space, $S$, into subsets, $S_k$, defined to contain all states that can violate a forbidding constraint after a minimum of $k$ state transitions. In other words, $S_0$ contains those states that violate a forbidding constraint directly; $S_1$ contains those states that do not violate any forbidding constraints themselves, but have a successor state (following the transition probability function, $P$) that does (i.e., a successor state in $S_0$); $S_2$ contains those states that do not violate any forbidding constraints, nor have any successors that do, but who have at least one successor state that has a successor state that does (i.e., a successor state in $S_1$); etc. There are at most $|S|$ nonempty subsets in this mutually exclusive sequence. To make this partition exhaustive, the special subset, $S_\infty$, contains all states from which the agent will never violate a forbidding constraint by following $P$. We first show, by induction over $k$, that $\forall s \in S_k$ $(0 \leq k \leq |S|)$, $V(s) = \langle t, N, U \rangle$, as required by the theorem.

**Basis step $(S_0)$:** By definition, the agent will violate a forbidding constraint in $s \in S_0$. Therefore, either $\exists c \in C_{fs}$ such that $c(s) = t$ or $\exists c \in C_{fa}$ such that $c(s, P(s)) = t$, so we know, from Equation 20, $V(s) = \langle t, N, U \rangle$.

**Inductive step $(S_k, 1 \leq k \leq |S|)$:** Assume, as the induction hypothesis, that $\forall s' \in S_{k-1}$, $V(s') = \langle t, N', U' \rangle$. By the definition of $S_k$, each state, $s \in S_k$, has at least one successor state, $s' \in S_{k-1}$. Then, according to Equation 20, $V(s) = \langle t, N, U \rangle$, because the disjunction over $S'$ must include $s'$, for which $F' = t$.

Therefore, by induction, we know that for all $s \in S_k$ $(0 \leq k \leq |S|)$, $V(s) = \langle t, N, U \rangle$. We now show that $\forall s \in S_\infty$, $V(s) = \langle f, N, U \rangle$. We prove, by induction over $t$, that, for any state, $s \in S_\infty$, $V^t(s) = \langle f, N, U \rangle$.

**Basis step $(V^0)$:** By definition, if $s \in S_\infty$, there cannot exist any $c \in C_{fs}$ such that $c(s) = t$. Then, from Equation 19, $V^0(s) = \langle f, N^0, U^0 \rangle$.

**Inductive step $(V^t, t > 0)$:** Assume, as the inductive hypothesis, that, for any $s' \in S_\infty$, $V^{t-1}(s') = \langle f, N', U' \rangle$. We know that $V^t(s) = \langle f, N^t, U^t \rangle$ if and only if all three disjunctions in Equation 20 are false. The first is false, as described in the basis step. The second term is similarly false, since, by the definition of $S_\infty$, there cannot exist any $c \in C_{fa}$ such that $c(s, P(s)) = t$. In evaluating the third term, we first note that $S' \subseteq S_\infty$. In other words, all of the successor states of $s$ are also in $S_\infty$ (if successor $s' \in S_k$ for some finite $k$, then $s \in S_{k+1}$). Since all of the successors are in $S_\infty$, we know, by the inductive hypothesis, that the disjunction over $V^{t-1}$ in all these successors is false. Therefore, all three disjunctive terms in Equation 20 are false, so $V^t(s) = \langle f, N^t, U^t \rangle$.

Therefore, by induction, we know that for all $s \in S_\infty$, $V(s) = \langle f, N, U \rangle$. By the definition of the state partition, these two results prove the theorem as required. $\square$

**Theorem 2** *An agent following policy, P, with value function, V, generated as described in Section 4.4, from any state $s \in S$ will satisfy each and every requiring constraint with probability one if and only if $V(s) = \langle F, C_{ra} \cup C_{rs}, U \rangle$ (for some U and F).*

**Proof Sketch:** The proof parallels that of Theorem 1, but with a state partition, $S_k$, where $k$ corresponds to the *maximum* number of transitions before satisfying a requiring constraint. However, here, states in $S_\infty$ are those that *violate* the constraint, rather than satisfy it. Some cycles in the state space can prevent a guarantee of satisfying a requiring constraint within any fixed number of transitions, although the probability of satisfaction *in the limit* may be 1. In our current constraint semantics, we have decided that such a situation fails to satisfy the constraint, and our algorithm behaves accordingly. Such cycles have no effect on the handling of forbidding constraints, where, as we saw for Theorem 1, we need consider only the *minimum*-length trajectory. □

The proofs of the two theorems operate independently, so the policy-specified action will satisfy all constraints, if such an action exists. The precedence of forbidding constraints over requiring ones has no effect on the optimal action in such states. However, if there are conflicting forbidding and requiring constraints in a state, then the preference ordering causes the agent to choose a policy that satisfies the forbidding constraint and violates a requiring constraint. The agent can make the opposite choice if we simply change the preference ordering from Section 4.4. Regardless of the choice, from Theorems 1 and 2, the agent can use the value function, $V$, to identify the existence of any such violation and notify the user of the violation and possible constraint conflict.